\documentclass{article}

\PassOptionsToPackage{numbers, compress}{natbib}

\usepackage[preprint]{neurips_2025}

\usepackage[utf8]{inputenc} 
\usepackage[T1]{fontenc}    
\usepackage{hyperref}       
\usepackage{url}            
\usepackage{booktabs}       
\usepackage{amsfonts}       
\usepackage{nicefrac}       
\usepackage{microtype}      
\usepackage{xcolor}         
\usepackage{mathtools}
\usepackage[inline]{enumitem}
\usepackage{graphicx}  
\usepackage[subrefformat=parens,labelformat=parens]{subcaption}
\usepackage{accents}
\usepackage{multirow, makecell}
\usepackage{pifont}
\usepackage{bm}
\usepackage{gensymb}

\DeclareMathOperator{\E}{\mathbb{E}}
\DeclareMathOperator{\tr}{tr}

\let\titleold\title
\renewcommand{\title}[1]{\titleold{#1}\newcommand{\thetitle}{#1}}

\newcommand{\first}[1]{\textbf{\textcolor[HTML]{B6321C}{#1}}}
\newcommand{\second}[1]{\textbf{\textcolor[HTML]{3C8031}{#1}}}
\newcommand{\third}[1]{\textbf{\textcolor[HTML]{006795}{#1}}}

\newcommand{\cmark}{\text{\ding{51}}}
\newcommand{\xmark}{\text{\ding{55}}}

\definecolor{myyellow}{HTML}{FFF000}
\definecolor{cyan}{HTML}{00FFFF}

\newcommand{\kbari}{\bar{k}_i}
\newcommand{\kstrokebari}{\bar{k}'_i}
\newcommand{\sigmabari}{\bar{\Sigma}_i}
\newcommand{\deltabari}{\bar{\delta}_i}

\newcommand{\pbari}{\bar{P}_i}
\newcommand{\pstrokebari}{\bar{P}'_i}
\newcommand{\kbaribeta}{{\bar{K}^{\beta}_i}}

\newcommand{\pbaribeta}{{\bar{P}^{\beta}_i}}

\newcommand{\betabari}[1][i]{\bar{\beta}_{#1}}

\newcommand{\rvkbari}{\mathbf{\bar{K}}_i}
\newcommand{\rvkbaribetai}{\mathbf{\bar{K}}^{\beta_i}_i}

\newcommand{\rvkbaribeta}{\mathbf{\bar{K}}^{\beta}_i}

\newcommand{\rvpbari}{\mathbf{\bar{P}}_i}
\newcommand{\rvpbaribetai}{\mathbf{\bar{P}}^{\beta_i}_i}
\newcommand{\rvpbaribeta}{{\mathbf{\bar{P}}^{\beta}_i}}

\newcommand{\rvhbari}{\mathbf{\bar{H}}_i}
\newcommand{\rvhbaribetai}{\mathbf{\bar{H}}^{\beta_i}_i}

\newcommand{\etabari}[1][i]{\bar{\eta}_{#1}}
\newcommand{\etabaribeta}[1][i]{\bar{\eta}^\beta_{#1}}

\newcommand{\lambdabari}{\bar{\Lambda}_i}

\newcommand{\distkbarinormal}{\mathcal{N}(\kbari, \sigmabari)}
\newcommand{\distkbaribeta}{p(\{\kbaribeta_l\}^{L_i}_{l=1}|\beta)}

\newcommand{\ktildeibeta}{{\tilde{K}^{\beta}_i}}
\newcommand{\ptildeibeta}{{\tilde{P}^{\beta}_i}}

\newcommand{\kcheckicand}{{\check{k}^{\mathit{cand}}_i}}
\newcommand{\kchecki}{{\check{k}_i}}
\newcommand{\deltakchecki}{\Delta \check{k}_i}

\newcommand{\ptildei}{\tilde{P}_i}
\newcommand{\scheck}{\check{S}_i}

\newcommand{\rvktildeibeta}{\mathbf{\tilde{K}}^\beta_i}

\newcommand{\rvptildeibeta}{{\mathbf{\tilde{P}}^{\beta}_i}}

\newcommand{\rvhtildeibeta}{{\mathbf{\tilde{H}}^{\beta}_i}}

\newcommand{\etatildeibeta}{\tilde{\eta}^\beta_i}

\newcommand{\lambdatildei}{\tilde{\Lambda}_i}

\newcommand{\distktildeibeta}{p(\{\ktildeibeta_l\}^{L_i}_{l=1}|\beta)}

\newcommand{\distptildeibeta}{p(\ptildei|\beta)}

\newcommand{\rvz}{{\mathbf{Z}}}

\newcommand{\etahatibeta}{\hat{\eta}^\beta_i}

\newcommand{\tsalient}{t_{\mathit{salient}}}
\newcommand{\tnoise}{t_{\mathit{noise}}}

\newcommand{\etatildebetamax}{\tilde{\eta}^\beta_{\mathit{max}}}

\newcommand{\lambdahati}{\hat{\Lambda}_i}

\title{Good Keypoints for the Two-View Geometry Estimation Problem}

\author{Konstantin Pakulev\textsuperscript{1}
\and
Alexander Vakhitov\textsuperscript{2}
\and
Gonzalo Ferrer\textsuperscript{1}
\and
\small{\textsuperscript{1}Skolkovo Institute of Science and Technology \hspace{20pt} \textsuperscript{2}Slamcore} \\ 
\small{\hypersetup{urlcolor=magenta}\urlstyle{same}\url{https://github.com/KonstantinPakulev/BoNeSS-ST}}
}

\begin{document}

\maketitle

\begin{abstract}

Local features are essential to many modern downstream applications. Therefore, it is of interest to determine the properties of local features that contribute to the downstream performance for a better design of feature detectors and descriptors. In our work, we propose a new theoretical model for scoring feature points (keypoints) in the context of the two-view geometry estimation problem. The model determines two properties that a good keypoint for solving the homography estimation problem should have: be repeatable and have a small expected measurement error. This result provides key insights into why maximizing the number of correspondences doesn't always lead to better homography estimation accuracy. We use the developed model to design a method that detects keypoints that benefit the homography estimation and introduce the \textbf{Bo}unded NeSS-ST (BoNeSS-ST) keypoint detector. The novelty of BoNeSS-ST comes from  \begin{enumerate*}[label=(\roman*)] \item strong theoretical foundations, \item a more accurate keypoint scoring due to subpixel refinement and \item a cost designed for superior robustness to low saliency keypoints\end{enumerate*}. As a result, BoNeSS-ST outperforms prior self-supervised local feature detectors on the planar homography estimation task and is on par with them on the epipolar geometry estimation task.

\end{abstract}

\section{Introduction}
\label{sec:intro}

Establishing image-to-image correspondences is an essential component of modern Structure-from-Motion~\cite{snavely2011scene, wu2013towards, schonberger2016structure} (SfM) and visual SLAM~\cite{klein2007parallel, forster2014svo, lim2014real, mur2015orb} (vSLAM) systems, which are used in areas of robotics and computer vision. Approaches for getting the correspondences between images can be classified into sparse, detector-free and dense. Sparse approaches~\cite{lowe2004distinctive,detone2018superpoint,tyszkiewicz2020disk} operate on images and produce a sparse set of features each of which is composed of a keypoint and its descriptor. Detector-free and dense approaches operate on pairs of images. Former~\cite{zhou2021patch2pix,sun2021loftr, truong2023topicfm} directly output correspondences by performing coarse matching of feature maps followed by pixel-level refinement. Latter~\cite{truong2021learning, edstedt2023dkm, edstedt2024roma} produce a dense warp by utilizing a 4D-correlation volume. 

Sparse approaches, due to their efficiency, present more interest for applications. Yet, to our knowledge, the influence of sparse features on downstream performance is scarcely studied. This problem poses great significance since if we have a model describing what features are good for downstream tasks it will enable designing better feature detectors and descriptors, thereby, impacting numerous applications depending on them. In our work, we study the relationship between feature points (keypoints) and the two-view geometry estimation problem, which is encountered during the initialization of SfM and vSLAM systems. We aim to answer the following question: what properties of keypoints make them good for the downstream task of the two-view geometry estimation? 

In Sec.~\ref{sec:expected_measurement_error} and Sec.~\ref{sec:stability_score}, we answer the posed question for the case when two views are related by a planar homography. In Sec.~\ref{sec:expected_measurement_error}, we demonstrate that the quality of a correspondence formed by a keypoint can be characterized by \textbf{the expected measurement error} (EME). In Sec.~\ref{sec:stability_score}, we extend this result to assessing the quality of a keypoint itself. Specifically, we devise an extension of EME called \textbf{\bm{$\beta$}-EME} that takes into account both the ability of the keypoint to form correspondences, i.e. \textbf{repeatability}, and the accuracy of formed correspondences, i.e. EME. We term $\beta$-EME in the score form as \textbf{the stability score}.  We conclude that a keypoint is good for solving the homography estimation problem if it has \underline{a small expected measurement error} and \underline{high repeatability}. We point out an important implication that follows from that conclusion: more correspondences, more inliers or better classical metrics do not guarantee to improve the accuracy of the homography estimation.

In Sec.~\ref{sec:boss}, we demonstrate that our mathematical model generalizes the notion of the stability score originally introduced in~\cite{pakulev2023ness} and reveals the prior formulation to be one of the upper bounds. Building on this result, we improve the training mechanism of the NeSS-ST keypoint detector~\cite{pakulev2023ness}, which uses the stability score for generating the supervisory signal. We contribute by \begin{enumerate*}[label=(\roman*)] \item deriving a tighter upper bound on the stability score (see Sec.~\ref{sec:boss}), \item designing a more accurate supervisory signal generation procedure leveraging sub-pixel refinement (see Sec.~\ref{sec:syss}) and \item proposing a new cost function with high robustness to low saliency keypoints for training the neural network (see Sec.~\ref{sec:boness}) \end{enumerate*}. We call our enhanced version of NeSS-ST trained in this manner \textbf{Bo}unded NeSS-ST (BoNeSS-ST).

In Sec.~\ref{sec:experiments}, we use BoNeSS-ST to experimentally verify our theoretical findings and show that they apply not only to the planar homography estimation problem but to the epipolar geometry estimation problem as well. BoNeSS-ST demonstrates downstream performance superior to its predecessor NeSS-ST~\cite{pakulev2023ness} and other self-supervised keypoint detectors~\cite{detone2018superpoint, barroso2019key, lee2022self} on HEB~\cite{barath2023large}, IMC-PT~\cite{jin2021image}, MegaDepth~\cite{li2018megadepth} and ScanNet~\cite{dai2017scannet}. BoNeSS-ST establishes a new state of the art on the homography estimation task among self-supervised keypoint detectors and shows competitive performance on the epipolar geometry estimation task.

\section{Related Work}
\label{sec:related_work}

How to determine the quality of feature points? A classical approach to this problem is to use metrics such as repeatability~\cite{schmid1998comparing}, matching score~\cite{mikolajczyk2005comparison} and mean matching accuracy~\cite{dusmanu2019d2}, which capture the proportion of correct correspondences. For this reason, most learned keypoint detectors are trained to maximize these metrics~\cite{detone2018superpoint, ono2018lf, revaud2019r2d2, barroso2019key, lee2022self} or the number of correspondences~\cite{tyszkiewicz2020disk}. However, a recent comprehensive study of local features~\cite{jin2021image} demonstrated that classical metrics do not guarantee a faithful assessment of a method's ability to perform well in downstream applications.

Another recent work~\cite{pakulev2023ness} provided further evidence that high classical metrics do not necessarily mean better downstream performance. The authors introduced a novel keypoint quality measure, the stability score, that assesses the robustness of a keypoint to viewpoint transformations. The work showed that detectors trained using that measure demonstrate excellent performance in the epipolar geometry estimation task but have poor classical metrics. 

GMM-IKRS~\cite{santellani2024gmm} is another work that explores the idea that repeatability alone is not a definitive measure of the quality of a keypoint. The authors suggest that apart from being repeatable a good keypoint should also have a small deviation, i.e. be detected close to the same position. The authors propose an interpretable score that allows to measure these properties via robustness and deviation.

In our work, we develop a mathematical model that generalizes the stability score~\cite{pakulev2023ness} and explains why it improves the downstream performance of keypoint detectors. We show that the stability score incorporates the properties of keypoints described by robustness and deviation~\cite{santellani2024gmm} via a different characterization. Our findings explain the discrepancy between the downstream performance on the two-view geometry estimation task and repeatability that was observed in~\cite{jin2021image}.

\section{Method}
\label{sec:method}

\subsection{Expected Measurement Error}
\label{sec:expected_measurement_error}

We start by considering the nonlinear refinement of a homography estimate between two images  $I_1$ and $I_2$ assuming given correspondences $\{k_i \xleftrightarrow{} \kstrokebari\}^n_{i=1}$. The keypoint $\kstrokebari$ on $I_2$ is known exactly. The keypoint $k_i$ is a noisy measurement of the keypoint $\kbari$ on $I_1$ such that $\kbari=\bar{H}\kstrokebari$,\footnote{Here and in the following, we denote the transformation of a vector $x$ by a homography $H$ as $Hx$ for readability.} where $\bar{H}$ is the homography between the views. If we assume that the measurement error of $\kbari$ obeys a Gaussian distribution $\mathcal{N}(\vec{0}, \sigmabari)$ then $k_i$ can be viewed as a realization of a random variable $\rvkbari \sim \distkbarinormal$. Under this assumption, the maximum likelihood (ML) estimate $\hat{H}$ of the homography $\bar{H}$ is obtained by minimizing the transfer error~\cite{hartley2003multiple} that is equivalent to maximizing the following log-likelihood:

\begin{equation}
    \label{eq:transfer_error_mle}
    \log f(\{k_i\}^{n}_{i=1} | H) = -\frac{1}{2}\sum^{n}_{i=1} \|k_i - H\kstrokebari\|^2_{\sigmabari},
\end{equation}

\noindent where $H$ is an initial homography estimate.

The ML estimate $\hat{H}$ is the optimal mapping (in terms of Eq.~\ref{eq:transfer_error_mle}) between noise-free keypoints $\{\kstrokebari\}^n_{i=1}$ and measured keypoints $\{k_i\}^n_{i=1}$, which, however, is not guaranteed to be a close approximation of the homography $\bar{H}$~\cite{hartley2003multiple}. For example, if $n = 4$ then $\hat{H}$ is the exact mapping of noise-free keypoints $\{\kstrokebari\}^n_{i=1}$ to measured keypoints $\{k_i\}^n_{i=1}$. Asymptotically, $\frac{1}{n}\sum^n_{i=1}\|\hat{H}\kstrokebari - \bar{k}_i\|^2_2 \xrightarrow{} 0$ as $n \xrightarrow{} \infty$. But, in practice, $n$ is limited by a keypoint budget; therefore, an estimate $\hat{H}$ that approximates $\bar{H}$ the best is achieved for the set $\{k_i \xleftrightarrow{} \kstrokebari\}^n_{i=1}$ with the smallest total measurement error $\sum^n_{i=1} \|\bar{H}\kstrokebari - k_i\|_2$. Thus, if we have a model of $\distkbarinormal$ then we can estimate $\bar{H}\kstrokebari$ and score correspondences benefitting the homography estimation according to individual $\|\bar{H}\kstrokebari - k_i\|_2$. In the following, we show that $\distkbarinormal$ can be formulated as a distribution of projections of noisy measurements of the same visual structure as $\kbari$ made in different views.

We define a noisy measurement (or a detection) as the application of a keypoint detector $D$ to a random image patch $\rvpbari \sim p(\pbari)$ depicting an area around the image structure that $\kbari$ corresponds to, where $p(\pbari)$ is a probability density function (PDF) of the distribution of such patches. Next, we consider a random homography $\rvhbari$ from $\rvpbari$ to $\pbari$, which is a function of $\rvpbari$ and $\pbari$, and define a random keypoint projection as $\rvhbari D(\rvpbari)$. One can notice that $\rvhbari D(\rvpbari)$, in general, follows a multimodal distribution since no keypoint detector $D$ is guaranteed to produce measurements related to $\kbari$ under arbitrary projective transformations. Therefore, we introduce a parameter $\beta_i$ that constrains the sample space of $\rvpbari$ based on the difficulty of viewpoint transformations. We denote by $\rvpbaribetai \sim p(\pbari|\beta_i)$ a random variable for which the sample space of corresponding $\rvhbaribetai$ is given as $\{\omega \in \Omega_{\rvhbari} | d(I, \rvhbari(\omega)) \leq \beta_i\}$, where $d$ is a function that measures the difference in difficulty between two homographies, yielding:

\begin{equation}
    \label{eq:rvkbaribetai}
    \rvkbaribetai = \rvhbaribetai D(\rvpbaribetai).
\end{equation}

\noindent Finally, we assume that $\rvkbaribetai=\rvkbari$ when $\beta_i=\betabari$ (see Fig.~\ref{fig:keypoint_projection_distribution}) such that $\forall \beta_i > \betabari$ $\rvkbaribetai$ follows a multimodal distribution (see Fig.~\ref{fig:beta}).

In practice, instead of noise-free keypoints $\{\kstrokebari\}^n_{i=1}$ we have access only to their noisy measurements $\{k'_i\}^n_{i=1}$, where $k'_i=D(\pstrokebari)$, and $\pstrokebari \in I_2$. So, we are interested in knowing $\|\bar{H}k'_i - k_i\|_2$ - not $\|\bar{H}\kstrokebari - k_i\|_2$. But, unlike $\bar{H}\kstrokebari$, which is a constant independent of the choice of the second view $I_2$, $\bar{H}k'_i$ does depend on $I_2$ since $k'_i$ is determined by the patch $\pstrokebari$. Therefore, we replace $\bar{H}k'_i$ by a random variable and obtain the following expectation:

\begin{equation}
    \label{eq:eme_bar_star}
    \etabari = \E_{\rvkbari \sim \distkbarinormal}[\|\rvkbari - k_i\|_2].
\end{equation}

\noindent We term the measure in Eq.~\ref{eq:eme_bar_star} \textbf{the expected measurement error} (EME). EME quantifies the instability of measurements of the image structure related to $\kbari$ and validates the fitness of a correspondence established by $k_i$ for the homography estimation task.

\subsection{Stability Score}
\label{sec:stability_score}

\begin{figure}
    \captionsetup[subfigure]{justification=centering}
    \begin{subfigure}{0.395\linewidth}
        \includegraphics[width=\linewidth]{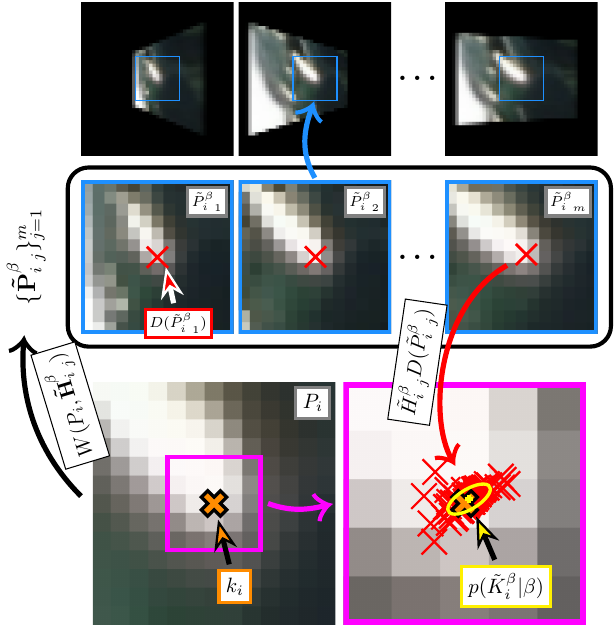}
        \caption{A unimodal $(L_i=1)$ distribution of keypoint projections.}
        \label{fig:keypoint_projection_distribution}
    \end{subfigure}
    \begin{subfigure}{0.6\linewidth}
        \includegraphics[width=\linewidth]{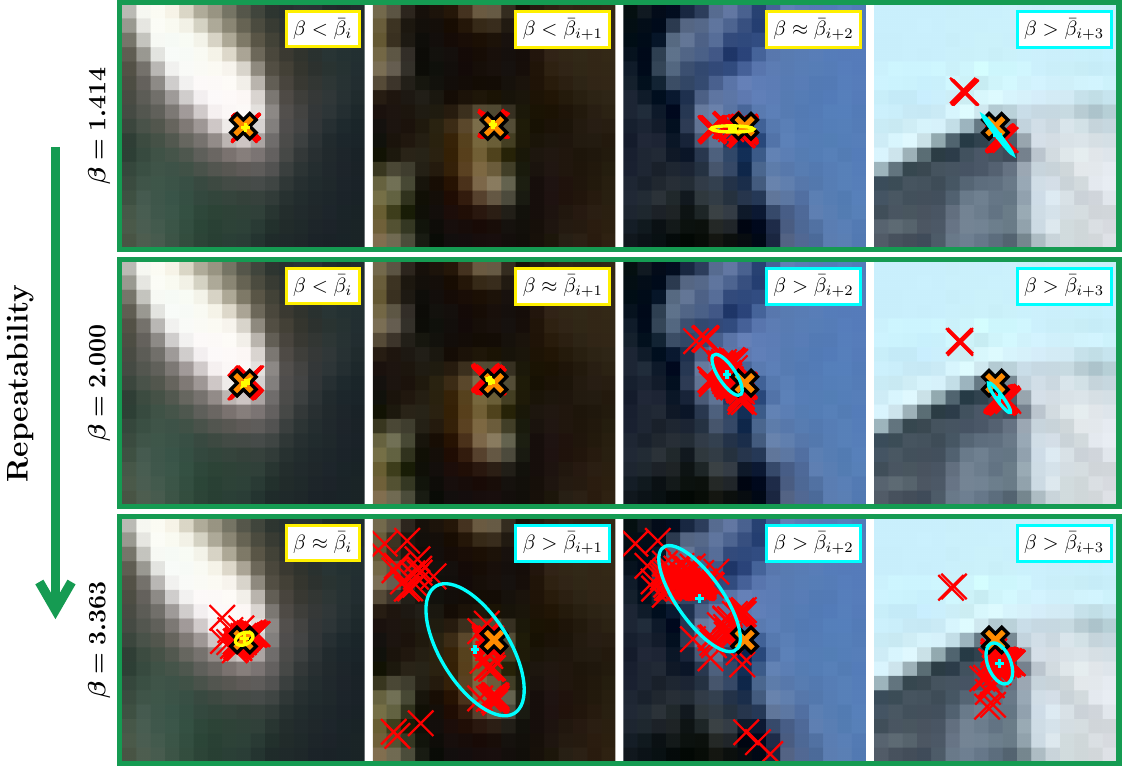}
        \caption{The influence of $\beta$ on the distribution of keypoint projections and repeatability.}
        \label{fig:beta}
    \end{subfigure}
    \caption{\textbf{\subref{fig:keypoint_projection_distribution}} We approximate $\distkbaribeta$ (see Sec.~\ref{sec:stability_score}) with $\distktildeibeta$ (see Sec.~\ref{sec:syss}) by using the distribution of synthetic patches with PDF $\distptildeibeta$. We generate random synthetic patches $\{\rvptildeibeta_j\}^m_{j=1}$ from $P_i$ via a function $W$ using random homographies $\{\rvhtildeibeta_j\}^m_{j=1}$ (see Appx.~\ref{appx:generation_of_synth_views}). \textbf{\subref{fig:beta}} For small values of $\beta$, $\distktildeibeta$ is unimodal for most keypoints and can be approximated by a Gaussian (\textsuperscript{\colorbox{myyellow}{}}). As $\beta$ increases, $\distktildeibeta$ can no longer be approximated well by a Gaussian (\textsuperscript{\colorbox{cyan}{}}) and remains unimodal only for repeatable keypoints. $\beta$-EME scores keypoints by both repeatability and EME, e.g $\etabari > \etabari[i+1]$ (see Eq.~\ref{eq:eme_bar_star}) but $\etabaribeta < \etabaribeta[i+1]$ for $\beta=3.363$ (see Eq.~\ref{eq:eme_bar_beta}).}
    \label{fig:keypoint_projection_distribution_and_beta}
\end{figure}


While EME allows for scoring correspondences, it is not a good criterion for scoring keypoints because a keypoint with a good EME may not have a sufficient \textbf{repeatability} to form a correspondence. Therefore, we modify EME to also account for the repeatability of a keypoint. We notice that the model of EME does have a component that characterizes repeatability - $\betabari$. A keypoint $\kbari$ with a larger $\betabari$ is measured by $D$ under more difficult transformations and, hence, is more repeatable (see Fig.~\ref{fig:beta}). Yet, EME doesn't reflect how large $\betabari$ is: given two keypoints with the same EME it is not possible to tell which one is more repeatable. Thus, to account for both repeatability and EME in a single measure, we take the expectation in Eq.~\ref{eq:eme_bar_star} using a distribution of viewpoint transformations that is shared between all keypoints, which is achieved by letting $\beta_i=\beta$ :

\begin{equation}
    \label{eq:eme_bar_beta}
    \etabaribeta = \E_{\rvkbaribeta \sim \distkbaribeta} [\|\rvkbaribeta - k_i\|_2].
\end{equation}

We refer to the measure in Eq.~\ref{eq:eme_bar_beta} as \textbf{\bm{$\beta$}-EME}. 
$\beta$-EME assesses the expected measurement error of a keypoint assuming apriori information $\beta$ about the difficulty of possible viewpoint transformations between the two views. In practice, it is useful to represent $\beta$-EME in the form a score, which we call \textbf{the stability score}, defined on a half-open interval $(0, 1]$ as:

\begin{equation}
    \label{eq:stability_score_bar_i_beta}
    \lambdabari = e^{-\etabaribeta}.
\end{equation}

Each $\rvkbaribeta$ is assigned by $\beta$-EME into one of the two categories: \begin{enumerate*}[label=(\roman*)]
    \item random variables following a unimodal distribution ($L_i=1$ and $\beta \leq \betabari$)  and
    \item random variables following a multimodal distribution ($L_i>1$ and $\beta > \betabari$)
\end{enumerate*}. If a keypoint falls into the first category then it meets the repeatability requirements set by $\beta$, and $\etabaribeta$ is a faithful EME of $k_i$ for the specified range of viewpoint transformations. Otherwise, the keypoint belongs to the second category and doesn't have the required repeatability from which it follows that $\etabaribeta > \etabari$ because $\etabaribeta$ is the expectation over a multimodal distribution. Thus, $\beta$-EME mixes the contributions of keypoint projections related to different image structures when $\beta$ is large (see Fig.~\ref{fig:beta}). So, as $\beta$ grows, $\beta$-EME gradually loses the ability to faithfully represent EME for some keypoints (see Sec.~\ref{sec:intrinsic_property}). This can be of a concern for a wide-baseline scenario, where using the largest possible $\beta$ is desirable (see Appx.~\ref{appx:limitations:stability_score}). Therefore, a careful tuning of $\beta$ is required (see Appx.~\ref{appx:ablation_study}).

An important implication from the developed model is that having more correspondences, which generally means a higher number of inliers and better classical metrics,  doesn't have to improve the accuracy of the homography estimation. According to our model, \textbf{a good keypoint} for solving the homography estimation problem \textbf{has to satisfy two properties}: \underline{be repeatable} and \underline{have a small expected measurement error}. We confirm this implication experimentally in Sec.~\ref{sec:experiments} by using a keypoint detector enhanced with the stability score (see Sec.~\ref{sec:boness}) and show that the implication also applies to the epipolar geometry estimation problem.

\subsection{Bounded Stability Score}
\label{sec:boss}

We now demonstrate that our model generalizes the notion of the stability score from~\cite{pakulev2023ness}. Consider the following relation obtained by applying Jensen's inequality to Eq.~\ref{eq:eme_bar_star}:

\begin{equation}
\label{eq:eme_jensen_x2}
\E[\|\rvkbari - k_i\|_2]^2 \leq \E[\|\rvkbari - k_i\|^2_2].
\end{equation}

\noindent When $\E[\|\rvkbari - k_i\|_2] \geq 1$, it holds that  $\E[\|\rvkbari - k_i\|_2] \leq \E[\|\rvkbari - k_i\|_2]^2$, and an upper bound on $\E[\|\rvkbari - k_i\|_2]$ writes as:

\begin{equation}
\label{eq:eme_bound_jensen_x2}
\E[\|\rvkbari - k_i\|_2] \leq \E[\|\rvkbari - k_i\|^2_2].
\end{equation}

\noindent The right-hand side of Eq.~\ref{eq:eme_bound_jensen_x2} can be further decomposed as:

\begin{equation}
\label{eq:eme_jensen_x2_rhs_decomposition}
\begin{split}
\E_{\rvkbari \sim \distkbarinormal}[\|\rvkbari - k_i\|^2_2] & = \E[\|\rvkbari - (\kbari + \delta_i)\|^2_2] \\ & = \E[\|\rvkbari - \kbari\|^2_2] + \deltabari^{\top}\deltabari \\
& =  \tr(\sigmabari) + \deltabari^{\top}\deltabari,
\end{split}
\end{equation}

\noindent where $\deltabari = k_i - \kbari$ is a constant. Then, assuming $k_i=\kbari$, the original formulation of the stability score~\cite{pakulev2023ness} corresponds to an upper bound on Eq.~\ref{eq:eme_bar_star}:

\begin{equation}
\label{eq:eme_bound_jensen_x2_rhs_coarse_decomposition}
\E[\|\rvkbari - k_i\|_2] \leq \tr(\sigmabari) \leq 2\|\sigmabari\|_2.
\end{equation}

For practical purposes (see Sec.~\ref{sec:boness} and Appx.~\ref{appx:ablation_study}), it is useful to have a bound that is exempt from limitations of Eq.~\ref{eq:eme_bound_jensen_x2}, which requires $\E[\|\rvkbari - k_i\|_2] \geq 1$, and Eq.~\ref{eq:eme_bound_jensen_x2_rhs_coarse_decomposition}, which additionally assumes $k_i=\kbari$ (the equality doesn't generally hold). Therefore, we further improve the bound in Eq.~\ref{eq:eme_bound_jensen_x2} by applying the monotonic function $\sqrt{x}$ to both sides of the inequality: 

\begin{equation}
\label{eq:eme_bound_jensen_sqrt_x2}
\E[\|\rvkbari - k_i\|_2] \leq \sqrt{\E[\|\rvkbari - k_i\|^2_2]}.
\end{equation}

\noindent Eq.~\ref{eq:eme_bound_jensen_sqrt_x2} holds for every $\E[\|\rvkbari - k_i\|_2]$ since $\E[\sqrt{x}] \leq \sqrt{\E[x]}$ from Jensen's inequality. We refer to the stability score calculated using a bound on Eq.~\ref{eq:eme_bar_beta} as {\bf the bounded stability score}.

Eq.~\ref{eq:eme_jensen_x2_rhs_decomposition} offers an interpretable decomposition of bounds in Eq.~\ref{eq:eme_bound_jensen_x2}, Eq.~\ref{eq:eme_bound_jensen_x2_rhs_coarse_decomposition} and Eq.~\ref{eq:eme_bound_jensen_sqrt_x2}. Concretely, the term $\tr(\sigmabari)$ is an upper bound on EME of $\kbari$ and characterizes the spread of measurements of the image structure related to $\kbari$. The term $\deltabari^{\top}\deltabari$ is a squared distance from $\kbari$ and characterizes the negative log-likelihood of $k_i$ to be close to a projection of a measurement, i.e. its likelihood to be repeated.

\begin{figure}
    \includegraphics[width=\linewidth]{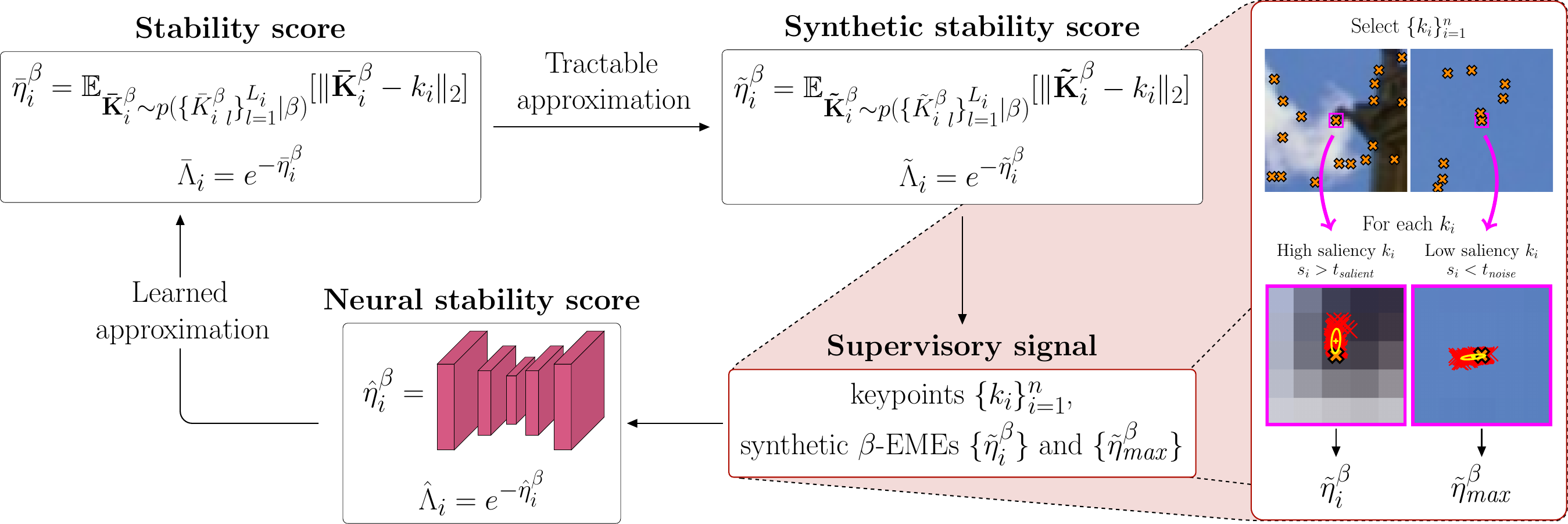}
    \caption{\textbf{Training pipeline.} Because the distribution of keypoints projections with PDF $\distkbaribeta$ is intractable, it is not possible to estimate the stability score $\lambdabari$ for a keypoint $k_i$. Therefore, we learn an approximation of $\beta$-EME (denoted as $\etahatibeta$) using the synthetic $\beta$-EME (denoted as $\etatildeibeta$) to create the supervisory signal. $\etatildeibeta$ constitutes only a coarse approximation of $\etabaribeta$ and may yield erroneous estimates, e.g for keypoints with low saliency $s_i < \tnoise$. We replace such estimates with $\etatildebetamax$ utilizing the knowledge that they are highly unlikely to be repeated.}
    \label{fig:boness_st_training}
\end{figure}

\subsection{Synthetic Stability Score}
\label{sec:syss}

Estimating the stability score or its bounds requires \begin{enumerate*}[label=(\roman*)]
      \item drawing $m$ samples $\{\rvpbaribeta_j\}^m_{j=1} \overset{\mathrm{i.i.d.}}{\sim} p(\pbari|\beta)$ and 
    \item producing measurements $\{D(\pbaribeta_j)\}^m_{j=1}$ using realizations $\{\pbaribeta_j\}^m_{j=1}$ of these samples.
\end{enumerate*} However, due to limitations of existing datasets with reconstructed correspondences, it is not possible to obtain samples of needed quality and in required quantities solely from real-world data~\cite{pakulev2023ness}. Therefore, we adopt an approximation of Eq.~\ref{eq:eme_bar_beta} using a distribution of synthetic patches with PDF $\distptildeibeta$, which we define via a procedure from~\cite{pakulev2023ness, detone2018superpoint} (see Fig.~\ref{fig:keypoint_projection_distribution} and Appx.~\ref{appx:generation_of_synth_views}):

\begin{equation}
    \label{eq:eme_tilde_beta}
    \etatildeibeta = \E_{\rvktildeibeta \sim \distktildeibeta} [\|\rvktildeibeta - k_i\|_2]. 
\end{equation}

\noindent We denote the stability score calculated via Eq.~\ref{eq:eme_tilde_beta} as $\lambdatildei$ calling it \textbf{the synthetic stability score}.

To produce a measurement $D(\ptildeibeta)$ for the realization $\ptildeibeta$ of a sampled synthetic view $\rvptildeibeta \sim \distptildeibeta$, we propose the following procedure based on~\cite{pakulev2023ness}. First, we apply the Shi-Tomasi detector~\cite{shi1994good} to $\ptildeibeta$ getting a score patch $\scheck$. Then, we apply non-maximum suppression to $\scheck$ and produce a candidate keypoint $\kcheckicand$ by selecting the pixel with the maximum score from $\scheck$. Next, we further refine $\kcheckicand$ by assuming a continuous second-order approximation $\scheck(\kchecki)$ of the Shi-Tomasi score $\scheck$ around $\kcheckicand$~\cite{pakulev2023ness, lowe2004distinctive}:

\begin{equation}
\label{eq:shi_tomasi_second_order_approximation}
\deltakchecki = -{\frac{\partial^2 \scheck}{\partial \kchecki^2}}^{-1}\frac{\partial \scheck}{\partial \kchecki}.
\end{equation}

\noindent If $\det \frac{\partial^2 \scheck}{\partial \kchecki^2} = 0$ or the condition number $\kappa(\frac{\partial^2 \scheck}{\partial \kchecki^2}) \gg 1$ then a meaningful solution to Eq.~\ref{eq:shi_tomasi_second_order_approximation} does not exist. Likewise, after applying non-maximum suppression, there might not be a suitable $\kcheckicand$. Such image structures for which it is not possible to obtain reliable measurements are not likely to be repeated in practice. Therefore, for $\etatildeibeta$ to be a better approximation of $\etabaribeta$, we output a measurement $c$ with a large measurement error for these visual structures instead of relying on Shi-Tomasi. Thus, the complete measurement procedure is as follows:

\begin{equation}
\label{eq:shi_tomasi_measurement}
D(\ptildeibeta)=\left\{
    \begin{array}{ll}
        \multirow{2}{*}{$\kcheckicand + \deltakchecki$} & \det \frac{\partial^2 \scheck}{\partial \kchecki^2} \neq 0 \hspace{4pt} \text{and}\\ & |\deltakchecki|_{\infty} < 0.5\\[4pt]
        c & \text{otherwise} \\
    \end{array}
    \right..
\end{equation}

\noindent We demonstrate the importance of Eq.~\ref{eq:shi_tomasi_measurement} in Appx.~\ref{appx:ablation_study}.

The distribution with PDF $p(\ptildei|\beta)$ features only viewpoint changes of the neighborhood of $\kbari$, so $\etatildeibeta$, the synthetic $\beta$-EME, can be vastly different from the true value $\etabaribeta$ for some of the keypoints (see Fig.~\ref{fig:boness_st_training}). This flaw limits using the synthetic stability score in practice.

\subsection{Bounded Neural Stability Score}
\label{sec:boness}

To overcome the limitations of the synthetic stability score, the authors in~\cite{pakulev2023ness} use $\etatildeibeta$ as a part of the supervisory signal to learn a more accurate approximation $\etahatibeta$ of $\etabaribeta$ from the data via a neural network (see Fig.~\ref{fig:boness_st_training}).\footnote{Since our approach generalizes the one proposed in~\cite{pakulev2023ness}, we use our notation for explaining NeSS to provide a self-contained narrative.} Hence, the stability score $\lambdahati$ calculated using $\etahatibeta$ is referred to as \textbf{the neural stability score} (NeSS). The neural approximation $\etahatibeta$ is learned by exploiting the observation that an erroneous $\etatildeibeta$ is highly likely to come from a low-saliency area. So, a keypoint $k_i$ with a Shi-Tomasi score $s_i$ is used as the ground truth for learning $\etahatibeta$ only if $s_i > \tsalient$, where $\tsalient$ is a threshold. Thus, this approach doesn't penalize keypoints with low saliency directly because some of them do correspond to meaningful signals. Instead, the approach relies on the structure of the neural network, which impedes fitting the parameters to noise~\cite{ulyanov2018deep}, to eliminate erroneous predictions for keypoints with low saliency corresponding to noisy image structures.

Based on our mathematical model (see Sec.~\ref{sec:stability_score}), we propose an improvement to the supervisory signal generation procedure for learning NeSS. Consider a keypoint with extremely low saliency such that $\etabari \approx 0$, e.g image noise. For this keypoint, small changes in the realization of $\rvpbaribeta$ yield measurements related to different image structures each of which is unlikely to be repeated. Therefore, when calculating a synthetic approximation of $\etabaribeta$ for such a keypoint, we can employ Eq.~\ref{eq:shi_tomasi_measurement} to return $c$ for each measurement. Then, $\etatildeibeta$ calculated from projections of these measurements equals $\etatildebetamax$ - the maximum possible value of $\beta$-EME. To get such low-saliency keypoints for training, we consider an extra threshold $\tnoise$ such that $\tnoise \ll \tsalient$ and use a keypoint $k_i$ as the ground truth if $s_i < \tnoise$. This allows us to provide ground truth for more keypoints compared to NeSS-ST~\cite{pakulev2023ness} and improve the training (see Fig.~\ref{fig:boness_st_training} and Appx.~\ref{appx:ablation_study}). 

The training of the neural network is organized as follows. Given an image, we use the Shi-Tomasi detector for obtaining positions of all keypoints for which $s_i > \tsalient \lor s_i < \tnoise$. Then, we employ the neural network with the so-far-trained weights~\cite{ono2018lf, barroso2019key, tyszkiewicz2020disk, pakulev2023ness} to predict $\etahatibeta$ for each pixel of the image and select $n$ keypoints $\{k_i\}^n_{i=1}$ among keypoints detected by Shi-Tomasi with the highest $\{\etahatibeta\}^n_{i=1}$. Next, we apply the bound from Eq.~\ref{eq:eme_bound_jensen_sqrt_x2} to Eq.~\ref{eq:eme_tilde_beta} and use it to estimate $\{\etatildeibeta\}^n_{i=1}$ for $\{k_i\}^n_{i=1}$. We process $\{\etatildeibeta\}^n_{i=1}$ to prepare the ground truth by replacing estimates for which $s_i < \tnoise$ with $\etatildebetamax$. Finally, we train the neural network using the following objective:

\begin{equation}
\label{eq:traning_loss}
    \begin{array}{ll}
        L = \frac{\sum\limits_{s_i > \tsalient}(\etahatibeta - \etatildeibeta)^2 + \sum\limits_{s_i < \tnoise}(\etahatibeta - \etatildebetamax)^2}{n}.
    \end{array}
\end{equation}

We call our detector BoNeSS-ST because it scores keypoints detected by Shi-Tomasi (ST) using \textbf{the bounded neural stability score} (BoNeSS), which is calculated using the inference of the neural network. Refer to Appx.~\ref{appx:training_details} and Appx.~\ref{appx:ablation_study} for more details about the training procedure and the ablation study of the influence of the proposed improvements on the performance of BoNeSS-ST.

\section{Experiments}
\label{sec:experiments}

We conduct the evaluation on the two-view geometry estimation task and classical metrics to support the statements of our theoretical model and highlight the importance of design choices made in Sec.~\ref{sec:method}. We compare BoNeSS-ST with Shi-Tomasi~\cite{shi1994good} and NeSS-ST~\cite{pakulev2023ness} for a better illustration of the advantages that a finer scoring of keypoints brings. Note, that all three methods rely on \textbf{the same keypoints}, which are provided by the Shi-Tomasi detector, but score them differently. The evaluation also includes other baseline and state-of-the-art methods: SIFT~\cite{lowe2004distinctive}, SuperPoint~\cite{detone2018superpoint}, R2D2~\cite{revaud2019r2d2}, Key.Net~\cite{barroso2019key}, DISK~\cite{tyszkiewicz2020disk}, REKD~\cite{lee2022self}.

\subsection{Evaluation on the Two-View Geometry Estimation Task}
\label{sec:two_view_geometry_eval}

We demonstrate that the stability score enhances the downstream performance of a keypoint detector on the homography estimation task as stated in Sec.~\ref{sec:stability_score} by evaluating on the planar homography estimation benchmark - HEB~\cite{barath2023large}. HEB is a relatively new dataset of landmark photos with a focus on planar surfaces. The dataset features strong viewpoint and illumination changes totaling 169k test image pairs with 4.8k unique test images. Additionally, we study how the results of Sec.~\ref{sec:stability_score} apply to the epipolar geometry estimation problem by evaluating on well-established benchmarks: IMC-PT~\cite{jin2021image}, MegaDepth~\cite{li2018megadepth} and ScanNet~\cite{dai2017scannet}.

A proper evaluation of keypoint detectors on downstream tasks requires using a single keypoint descriptor for all methods~\cite{mikolajczyk2005comparison, verdie2015tilde, zhang2017learning, barroso2019key, pakulev2023ness}. We use the DISK descriptor~\cite{tyszkiewicz2020disk} on IMC-PT~\cite{jin2021image} and MegaDepth~\cite{li2018megadepth} and the HardNet~\cite{mishchuk2017working} descriptor on ScanNet~\cite{dai2017scannet}. For HEB~\cite{barath2023large}, we also choose DISK due to its good performance. Another important component of the evaluation is choosing hyperparameters such as the ratio test threshold~\cite{lowe2004distinctive} and the inlier threshold for each method~\cite{jin2021image}. To find optimal hyperparameters, we build on a pipeline from~\cite{pakulev2023ness}, which is used for finding initial values of hyperparameters in a greedy manner~\cite{jin2021image}, and enhance it by the grid search approach suggested in~\cite{barath2023large}. Refer to Appx.~\ref{appx:eval_protocol} and Appx.~\ref{appx:limitations:evaluation_protocol} for more details about the evaluation protocol.

\paragraph{Results.} The evaluation of NeSS-ST and BoNeSS-ST detectors (see Table~\ref{tab:evaluation}) shows that replacing the criterion for scoring keypoints detected by Shi-Tomasi with the stability score yields significant improvements on planar homography estimation and epipolar geometry estimation tasks. Table~\ref{tab:evaluation} also demonstrates that the number of inliers is not a definitive measure of the quality of the two-view geometry estimation accuracy. The results indicate that apart from their numbers there is an equally important ``intrinsic'' property of inliers that influences mAA. In Sec.~\ref{sec:intrinsic_property}, we provide evidence that this intrinsic property is the expected measurement error. Thus, the presented experimental data align with the conclusion of Sec.~\ref{sec:stability_score} about the properties of keypoints that are good for the homography estimation problem and show that the same considerations apply to the epipolar geometry estimation problem. Refer to Appx.~\ref{appx:two_view_geometry_eval} for more evaluation results.

BoNeSS-ST establishes a new state of the art among self-supervised keypoint detectors on the homography estimation task by outperforming all its competitors: SuperPoint~\cite{detone2018superpoint}, Key.Net~\cite{barroso2019key}, REKD~\cite{lee2022self} and NeSS-ST~\cite{pakulev2023ness}. Our method demonstrates very good generalization: trained on MegaDepth, it performs well on all datasets ranking in top-3 on all of them. A good performance of Shi-Tomasi, SIFT, SuperPoint, NeSS-ST and BoNeSS-ST demonstrates that the classical definition of a keypoint~\cite{lindeberg1998feature} (a blob or a corner) is still highly competitive. For this reason, we believe that hybrid learned-classical approaches~\cite{barroso2019key, pakulev2023ness} and approaches inspired by the classical definition of a keypoint~\cite {detone2018superpoint} are a promising direction of research.

\begin{table}
    \caption{\textbf{Evaluation on the two-view geometry estimation task and with classical metrics.} We report relative pose rotation (R) and translation (t) mAA~\cite{yi2018learning, jin2021image} for a $10\degree$ threshold as well as the number of inliers (NI). Also, we report repeatability (Rep.)~\cite{schmid1998comparing} and MMA~\cite{mikolajczyk2005performance, dusmanu2019d2} for a 3-pixel threshold. We denote self-supervised methods by \cmark. The top-3 results in each column are in \first{red}, \second{green} and \third{blue}.}
    \label{tab:evaluation}
    \begin{center}
    \resizebox{\linewidth}{!}{
    \begin{tabular}{ccccccccccccccc}
    \toprule
    \multirow{2}{*}{Method} & \multirow{2}{*}{\makecell{Self-\\ sup.}} &
    \multicolumn{2}{c}{HEB~\cite{barath2023large}} &
    \multicolumn{3}{c}{IMC-PT~\cite{jin2021image}} &
    \multicolumn{3}{c}{MegaDepth~\cite{li2018megadepth}} &
    \multicolumn{3}{c}{ScanNet~\cite{dai2017scannet}} & 
    \multicolumn{2}{c}{HPatches~\cite{balntas2017hpatches}} \\
    
    & & R & NI & 
    R & t & NI & 
    R & t & NI & 
    R & t & NI &
    Rep. & MMA \\
    
    \midrule

    SIFT~\cite{lowe2004distinctive} & NA &
    0.182 & 49.2 & 
    0.714 & 0.390 & 181.3 & 
    0.844 & 0.292 & 383.6 & 
    0.606 & 0.211 & 110.9 &
    0.486 & 0.740 \\
    
    SuperPoint~\cite{detone2018superpoint} & $\cmark$ &
    0.228 & \third{57.5} & 
    0.726 & 0.383 & 137.1 & 
    0.873 & \third{0.321} & 405.5 & 
    0.643 & 0.234 & 107.4 &
    \first{0.617} & \third{0.759} \\
    
    R2D2~\cite{revaud2019r2d2} & \xmark &
    \second{0.232} & \second{66.0} &
    0.753 & 0.410 & \second{250.4} &
    0.873 & 0.312 & \second{452.2} &
    \first{0.669} & \first{0.251} & \first{167.6} &
    \second{0.608} & 0.748 \\
    
    Key.Net~\cite{barroso2019key} & $\cmark$ &
    0.201 & 54.3 &
    0.651 & 0.304 & 138.6 &
    0.840 & 0.263 & 384.6 &
    0.588 & 0.206 & 90.7 &
    \third{0.596} & 0.685 \\
    
    DISK~\cite{tyszkiewicz2020disk} & \xmark &
    \first{0.247} & \first{118.8} &
    \first{0.818} & \first{0.487} & \first{470.4} &
    \second{0.883} & 0.321 & \first{622.3} &
    0.537 & 0.165 & 138.4 &
    0.577 & \second{0.763} \\
    
    REKD~\cite{lee2022self} & $\cmark$ &
    0.216 & 55.1 &
    0.649 & 0.301 & 126.9 &
    0.855 & 0.274 & 394.6 &
    \second{0.657} & \second{0.241} & 102.7 &
    0.542 & 0.691 \\
    
    \midrule
        
    Shi-Tomasi~\cite{shi1994good} & NA &
    0.197 & 57.0 & 
    0.752 & 0.426 & \third{215.0} & 
    0.863 & 0.307 & \third{429.1} & 
    0.591 & 0.209 & \third{149.2} &
    0.579 & \first{0.772} \\
    
    NeSS-ST~\cite{pakulev2023ness} & $\cmark$ &
    0.228 & 42.9 &
    \third{0.760} & \third{0.434} & 148.9 &
    \third{0.876} & \second{0.333} & 286.8 &
    0.633 & 0.225 & 105.1 &
    0.456 & 0.702 \\
    
    BoNeSS-ST & $\cmark$ &
    \third{0.230} & 51.9 &
    \second{0.789} & \second{0.460} & 208.1 &
    \first{0.886} & \first{0.340} & 344.7 &
    \third{0.653} & \third{0.238} & \second{154.0} &
    0.504 & 0.745 \\

    \bottomrule
    \end{tabular}
    }
    \end{center}
\end{table}

\subsection{Evaluation on Classical Metrics}
\label{sec:classical_metrics_eval}

We continue illustrating the validity of the claims made in the conclusion of Sec.~\ref{sec:stability_score} by evaluating on the HPatches dataset~\cite{balntas2017hpatches} using classical metrics: repeatability~\cite{schmid1998comparing} and mean matching accuracy (MMA)~\cite{mikolajczyk2005performance, dusmanu2019d2}. We use the DISK descriptor~\cite{tyszkiewicz2020disk} for this dataset.

\paragraph{Results.} Table~\ref{tab:evaluation} demonstrates a major discrepancy in the ranking of the methods on downstream tasks and classical metrics. For instance, Shi-Tomasi is superior to NeSS-ST and BoNeSS-ST according to classical metrics but is inferior as far as mAA is concerned. Thus, not only the number of inliers but also the proportion of correct correspondences, which is characterized by classical metrics, is not a definitive measure of a method's ability to perform well on the two-view geometry estimation task. Nevertheless, similarly to the number of inliers, classical metrics can strongly correlate with the improvements in downstream performance, which can be seen by comparing the performance of BoNeSS-ST with NeSS-ST. This, again, hints at the existence of some intrinsic property of a keypoint, which we discuss in the following section. Refer to Appx.~\ref{appx:classical_metrics_eval} for more evaluation results.

\subsection{Intrinsic Property of a Keypoint}
\label{sec:intrinsic_property}

The experimental data in Sec.~\ref{sec:two_view_geometry_eval} and Sec.~\ref{sec:classical_metrics_eval} indicate that a keypoint possesses a certain intrinsic property that is necessary to account for to get accurate two-view geometry estimates. We provide conclusive evidence that the expected measurement error is that property. Concretely, we sweep the parameter $\beta$, which controls repeatability in our model (see Sec.~\ref{sec:stability_score}), and study the effect of changing it on the downstream performance and classical metrics on IMC-PT~\cite{jin2021image} and viewpoint sequences of HPatches~\cite{balntas2017hpatches}. For this experiment, the performance of NeSS-ST with $\beta=2.0$ and BoNeSS-ST with $\beta=2.828$ differs from the one reported in Table~\ref{tab:evaluation} (see Appx.~\ref{appx:generation_of_synth_views},  Appx.~\ref{appx:eval_protocol} and Appx.~\ref{appx:ablation_study} for details). We do not report the performance of NeSS-ST for $\beta=1.189$, $\beta=3.363$ and $\beta=4.0$ because the homography generation procedure used in~\cite{pakulev2023ness} is not able to generate homographies that conform to these values of $\beta$ (see Appx.~\ref{appx:generation_of_synth_views}).

\paragraph{Results.} 

Table~\ref{tab:beta_ablation} shows that although BoNeSS-ST with $\beta=1.189$ has the lowest repeatability among all methods on viewpoint sequences (see Appx.~\ref{appx:classical_metrics_eval}), it is still able to outperform every other method apart from DISK in relative pose mAA on IMC-PT by a noticeable margin (see Table~\ref{tab:evaluation}). Increasing $\beta$ improves the performance of BoNeSS-ST, but eventually $\beta$ becomes too high for $\beta$-EME to faithfully describe EME for all keypoints as explained in Sec.~\ref{sec:stability_score}. This makes the scoring priority shift to repeatable keypoints, which might not have the best EMEs, that leads to a decrease in the downstream performance. Changing $\beta$ from $1.189$ to $4.0$ significantly narrows the gap in repeatability between BoNeSS-ST and Shi-Tomasi, yet BoNeSS-ST is able to retain excellent relative pose mAA. This indicates that some repeatable keypoints detected by the Shi-Tomasi detector present a suboptimal choice for solving the epipolar geometry estimation problem (see Appx.~\ref{appx:qualitative_results}). Thus, we conclude that accounting for the expected measurement error is essential for having a good two-view geometry estimation performance, which supports the conclusion of Sec.~\ref{sec:stability_score}.

\begin{table}
    \caption{\textbf{Ablation study of the influence of the parameter \bm{$\beta$} on the performance.} We report relative pose rotation (R) and translation (t) mAA~\cite{yi2018learning, jin2021image} for a $10\degree$ threshold as well as the number of inliers (NI). Also, we report repeatability (Rep.)~\cite{schmid1998comparing} and MMA~\cite{mikolajczyk2005performance, dusmanu2019d2} for a 3-pixel threshold. The best results are in \first{red}.}
    \label{tab:beta_ablation}
    \begin{center}
    \resizebox{\linewidth}{!}{
    \begin{tabular}{ccccccccccc}
    \toprule
    \multirow{3}{*}{$\beta$} & \multicolumn{5}{c}{NeSS-ST~\cite{pakulev2023ness}} & \multicolumn{5}{c}{BoNeSS-ST} \\
    & \multicolumn{3}{c}{IMC-PT~\cite{jin2021image}} & \multicolumn{2}{c}{HPatches (view.)~\cite{balntas2017hpatches}} & \multicolumn{3}{c}{IMC-PT~\cite{jin2021image}} & \multicolumn{2}{c}{HPatches (view.)~\cite{balntas2017hpatches}} \\
    & 
    R & t & NI & Rep. & MMA & 
    R & t & NI & Rep. & MMA  \\
    
    \midrule

    1.189 &
    - & - & - & - & - &
    0.781 & 0.453 & 200.2 & 0.471 & 0.684 \\

    1.414 &
    0.760 & 0.434 & 156.1 & 0.441 & 0.671 &
    0.783 & 0.456 & 201.0 & 0.492 & 0.695 \\
    
    1.681 &
    0.763 & 0.438 & 163.9 & 0.457 & 0.675 &
    0.786 & 0.457 & 204.9 & 0.503 & 0.695 \\
    
    2.0 &
    \first{0.765} & \first{0.440} & 168.3 & 0.489 & 0.685 &
    0.788 & \first{0.459} & 211.1 & 0.529 & 0.703 \\
    
    2.378 &
    0.764 & 0.437 & 170.4 & 0.502 & 0.686 &
    \first{0.789} & 0.458 & \first{215.4} & 0.549 & 0.705 \\
    
    2.828 &
    0.763 & 0.439 & \first{172.1} & \first{0.567} & \first{0.700} &
    0.788 & 0.457 & 212.4 & 0.557 & 0.704 \\

    3.363 &
    - & - & - & - & - &
    0.786 & 0.453 & 212.2 & 0.565 & 0.709 \\

    4.0 &
    - & - & - & - & - &
    0.784 & 0.450 & 210.3 & \first{0.558} & \first{0.711} \\

    \bottomrule
    \end{tabular}
    }
    \end{center}
\end{table}
\section{Conclusion}

In this work, we developed a mathematical model that describes the properties that a good keypoint for the homography estimation problem should have: repeatability and expected measurement error. The main implication from the presented model is that having more correspondences, inliers or better classical metrics does not always lead to superior two-view geometry estimates. We experimentally verified this claim and demonstrated that accounting for the expected measurement error enables retaining good downstream performance for keypoints with different repeatability. Additionally, we showed that the developed theory generalizes the notion of the stability score presented in the prior work and used the theory to improve the training of the NeSS-ST detector~\cite{pakulev2023ness}. Titled BoNeSS-ST, our enhanced keypoint detector has good generalization and demonstrates state-of-the-art performance on multiple datasets and tasks.

{
    \small
    \bibliographystyle{ieeenat_fullname}
    \bibliography{main}
}

\appendix
\newpage

\begin{center}
  \Large
  \textbf{\thetitle}\\
  \vspace{0.5em}Supplementary Material \\
  \vspace{1.0em}
\end{center}

\section{Generation of Synthetic Views}
\label{appx:generation_of_synth_views}

\begin{figure}[h]
  \captionsetup[subfigure]{justification=centering}
  \begin{center}
  \begin{subfigure}[b]{0.3\linewidth}
    \includegraphics[width=\linewidth]{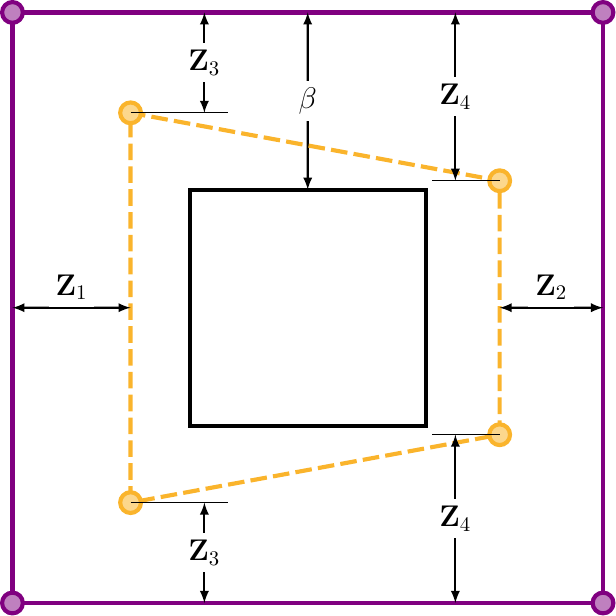}
    \caption{Homography generation procedure}
    \label{fig:homography_generation}
  \end{subfigure}
  \hfill
  \begin{subfigure}[b]{0.6\linewidth}
    \includegraphics[width=\linewidth]{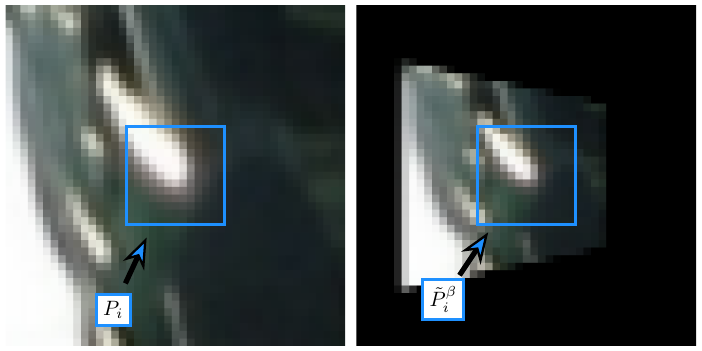}
    \caption{Effect of applying a generated homography to a part of an image.}
    \label{fig:patch_warping}
  \end{subfigure}
  \end{center}
  \caption{\textbf{\subref{fig:homography_generation}} Perturbed points (colored in yellow) are obtained by applying left, right and two perspective displacements, which are controlled by $\rvz_1$, $\rvz_2, \rvz_3$ and $\rvz_4$ respectively, to fixed points (colored in purple).  The parameter $\beta$ controls the magnitude of the displacements by preventing any of them from going inside the black square, which allows to regulate the difficulty of a generated homography. \textbf{\subref{fig:patch_warping}} Since we use the Shi-Tomasi detector, the size of $\ptildeibeta$, which is the realization of $\rvptildeibeta$, that is required to procedure a single measurement is rather small - $13 \times 13$ pixels. So, warped patches can be generated fast with a reasonable GPU memory budget, enabling online ground-truth generation during training.}
\end{figure}

We define the distribution with PDF $p(\ptildei|\beta)$ by warping $P_i$ with synthetic homographies, which are generated using a modified procedure from~\cite{pakulev2023ness, detone2018superpoint}. Our procedure outputs a random homography $\rvhtildeibeta$ by solving for the transformation between a set of 4 fixed points and a set of 4 perturbed points (see Fig.~\ref{fig:homography_generation}). The perturbed points are obtained by modifying the fixed points with four random variables sampled from the standard uniform distribution $\rvz_1, \rvz_2, \rvz_3, \rvz_4 \overset{\mathrm{iid}}{\sim} \mathcal{U}(0, 1)$. The difficulty of generated homographies is controlled by $\beta$ as illustrated in Fig.~\ref{fig:homography_generation}. Thus, denoting the procedure by $G^\beta$, a random homography $\rvhtildeibeta$ is obtained as:

\begin{equation}
  \label{eq:homography_generation}
  \rvhtildeibeta=G^\beta(\rvz_1, \rvz_2, \rvz_3, \rvz_4).
\end{equation}

\noindent Since not every homography is a plausible viewpoint transformation~\cite{detone2018superpoint}, the generated homographies are limited to perspective transformations, which are a special case of homographies. Given a random homography, a random patch $\rvptildeibeta$ is obtained by applying a warping function $W$ to the patch $P_i$:

\begin{equation}
\label{eq:patch_warping}
\rvptildeibeta = W(P_i, \rvhtildeibeta).
\end{equation}

\noindent See Fig.~\ref{fig:patch_warping} for an example of a warped patch.

Compared to the homography generation procedure from~\cite{pakulev2023ness}, we use two random variables for controlling left and right perspective displacements instead of a single one. This allows for a more fine-grained control of the difficulty of a generated homography and makes the difficulty grow more slowly as $\beta$ increases compared to the original procedure. For this reason, in Sec.~\ref{sec:intrinsic_property}, MMA and repeatability of NeSS-ST for which we use the original procedure grow faster with $\beta$ than that of BoNeSS-ST.

\section{Training Details}
\label{appx:training_details}

We train our models from scratch on MegaDepth~\cite{li2018megadepth} and validate them on IMC-PT~\cite{jin2021image} using subsets from~\cite{pakulev2023ness}. BoNeSS-ST employs the same U-Net architecture~\cite{ronneberger2015u} as NeSS-ST and, hence, retains its fast training and inference time. For training, we keep the hyperparameters and most of the parameters for generating the supervisory signal from NeSS-ST unchanged. We set the new parameter as $\tnoise=0.0001$ and use $\beta=2.828$ (see Appx.~\ref{appx:ablation_study} for the details about selecting $\beta$). For ablations, we train models belonging to the same ablation for an equal amount of time to avoid different training times affecting the results. The reference BoNeSS-ST model was trained in 19 hours on a single NVIDIA A100 GPU, which is comparable to the training time of NeSS that is 22 hours on NVIDIA 2080 TI~\cite{pakulev2023ness}.

\section{Evaluation Protocol}
\label{appx:eval_protocol}

\begin{figure}[h]
    \begin{center}
    \begin{subfigure}{\linewidth}
    \begin{center}
    \includegraphics[width=\linewidth]{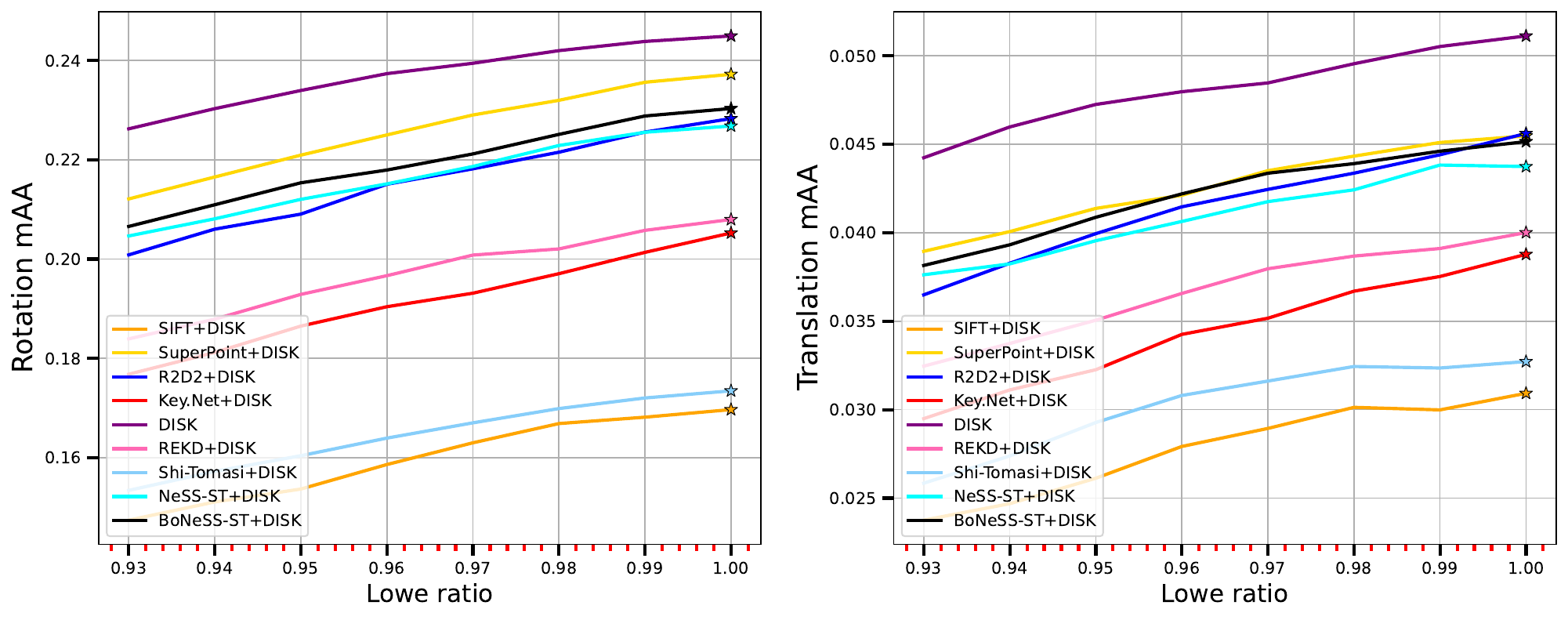}
      \caption{Ratio test threshold tuning.}
    \end{center}
    \end{subfigure}
    \begin{subfigure}{\linewidth}
    \begin{center}
    \includegraphics[width=\linewidth]{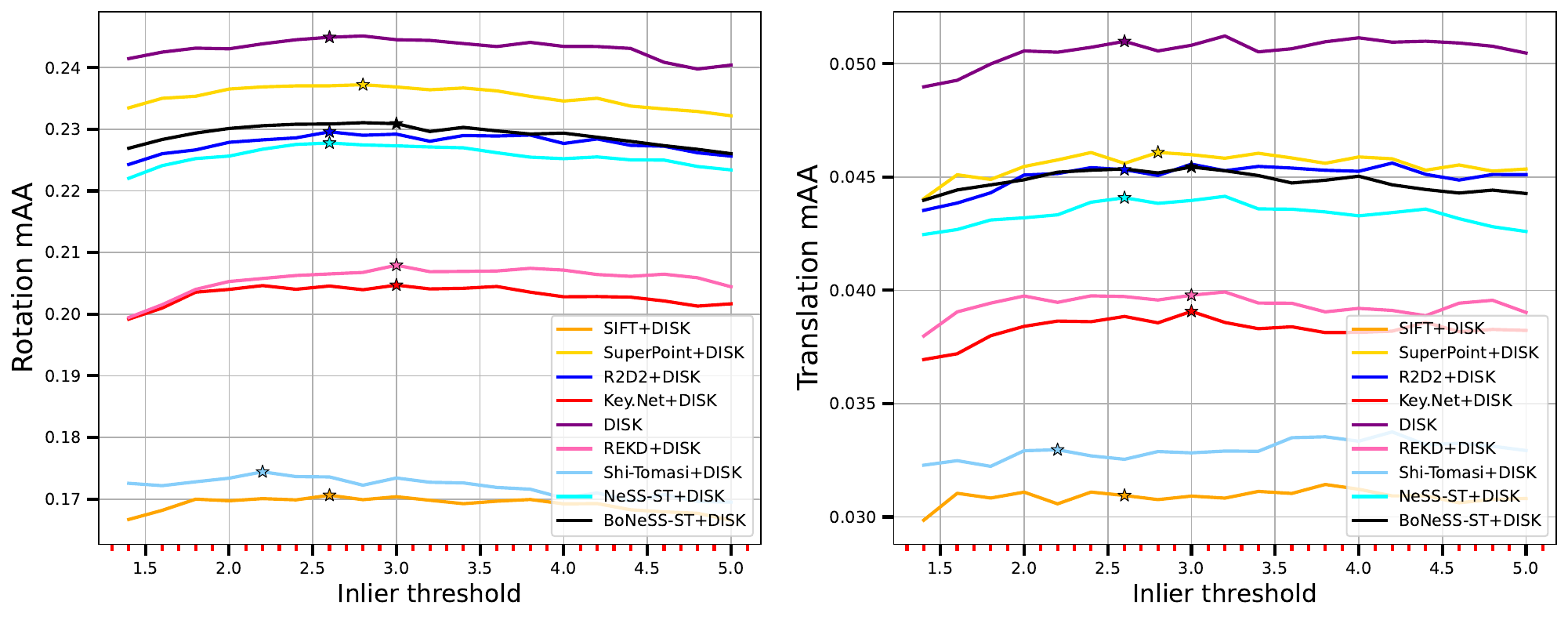}
    \caption{Inlier threshold tuning.}
    \end{center}
    \end{subfigure}
    \caption{\textbf{Initial tuning of method-specific hyperparameters for the HEB dataset.} We plot relative pose mAA-to-hyperparameter curves for rotation and translation calculated on the validation subset of HEB~\cite{barath2023large}. The star marks the best hyperparameter for a method according to the sum of rotation and translation mAA~\cite{yi2018learning, jin2021image} selected in a greedy manner~\cite{jin2021image}.}
    \label{fig:heb_hyperparams_tuning}
    \end{center}
\end{figure}

\begin{figure}[h]
    \begin{center}
    \begin{subfigure}{\linewidth}
    \begin{center}
    \includegraphics[width=\linewidth]{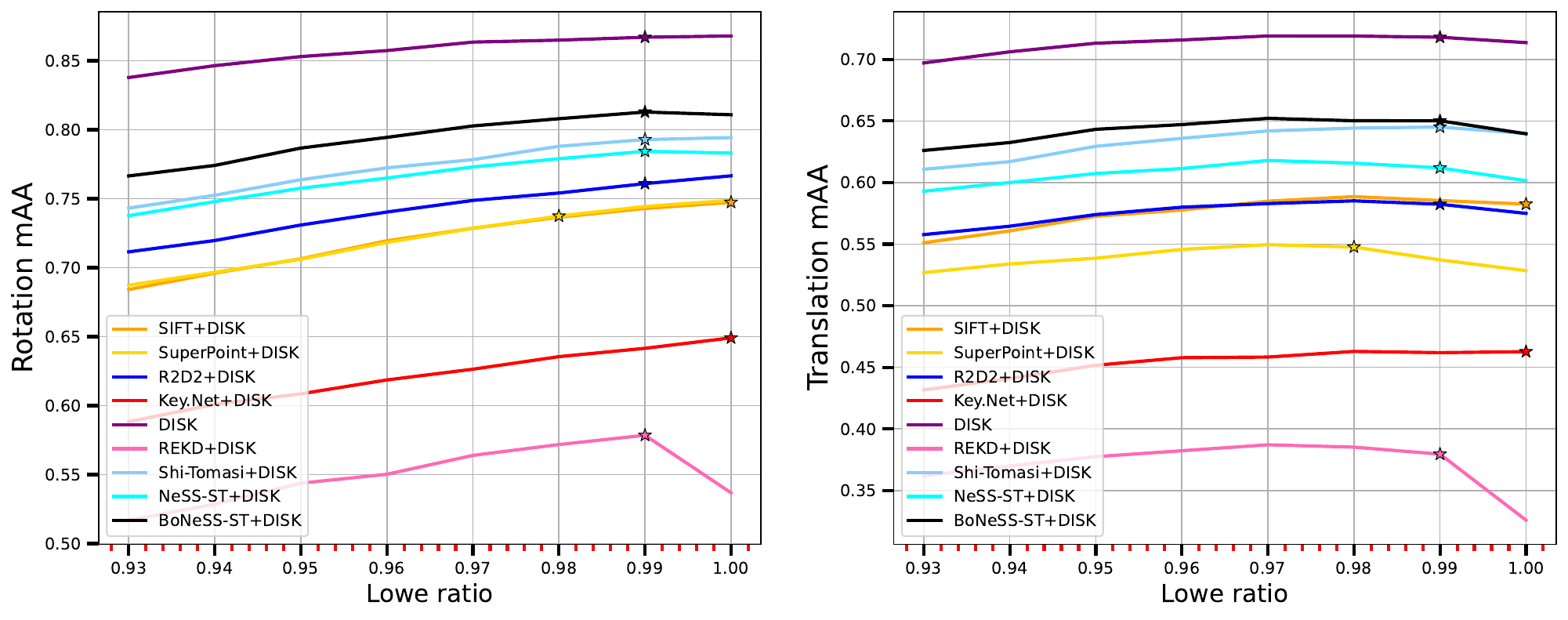}
      \caption{Ratio test threshold tuning.}
    \end{center}
    \end{subfigure}
    \begin{subfigure}{\linewidth}
    \begin{center}
    \includegraphics[width=\linewidth]{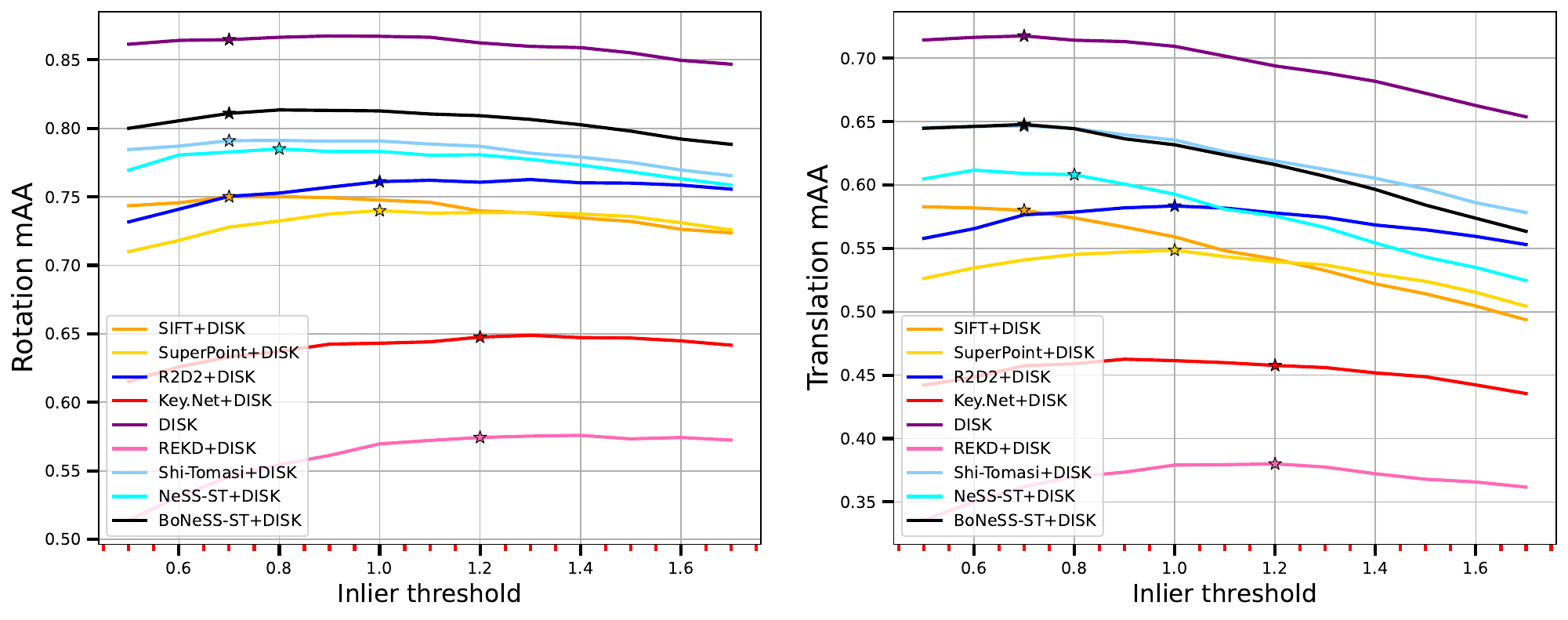}
    \caption{Inlier threshold tuning.}
    \end{center}
    \end{subfigure}
    \caption{\textbf{Initial tuning of method-specific hyperparameters for the IMC-PT dataset.} We plot relative pose mAA-to-hyperparameter curves for rotation and translation calculated on the validation subset of IMC-PT~\cite{jin2021image}. The star marks the best hyperparameter for a method according to the sum of rotation and translation mAA~\cite{yi2018learning, jin2021image} selected in a greedy manner~\cite{jin2021image}.}
    \label{fig:imc_pt_hyperparams_tuning}
    \end{center}
\end{figure}

\begin{figure}[h]
    \begin{center}
    \begin{subfigure}{\linewidth}
    \begin{center}
    \includegraphics[width=\linewidth]{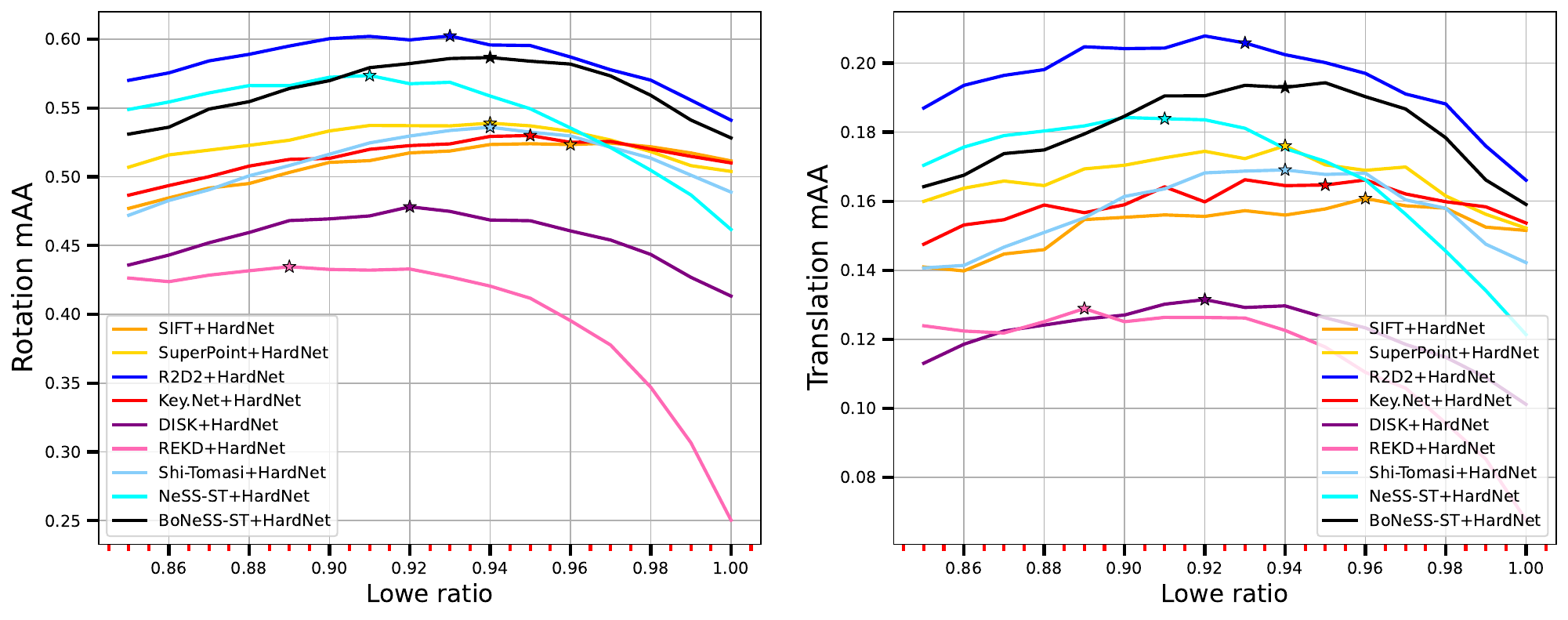}
      \caption{Ratio test threshold tuning.}
    \end{center}
    \end{subfigure}
    \begin{subfigure}{\linewidth}
    \begin{center}
    \includegraphics[width=\linewidth]{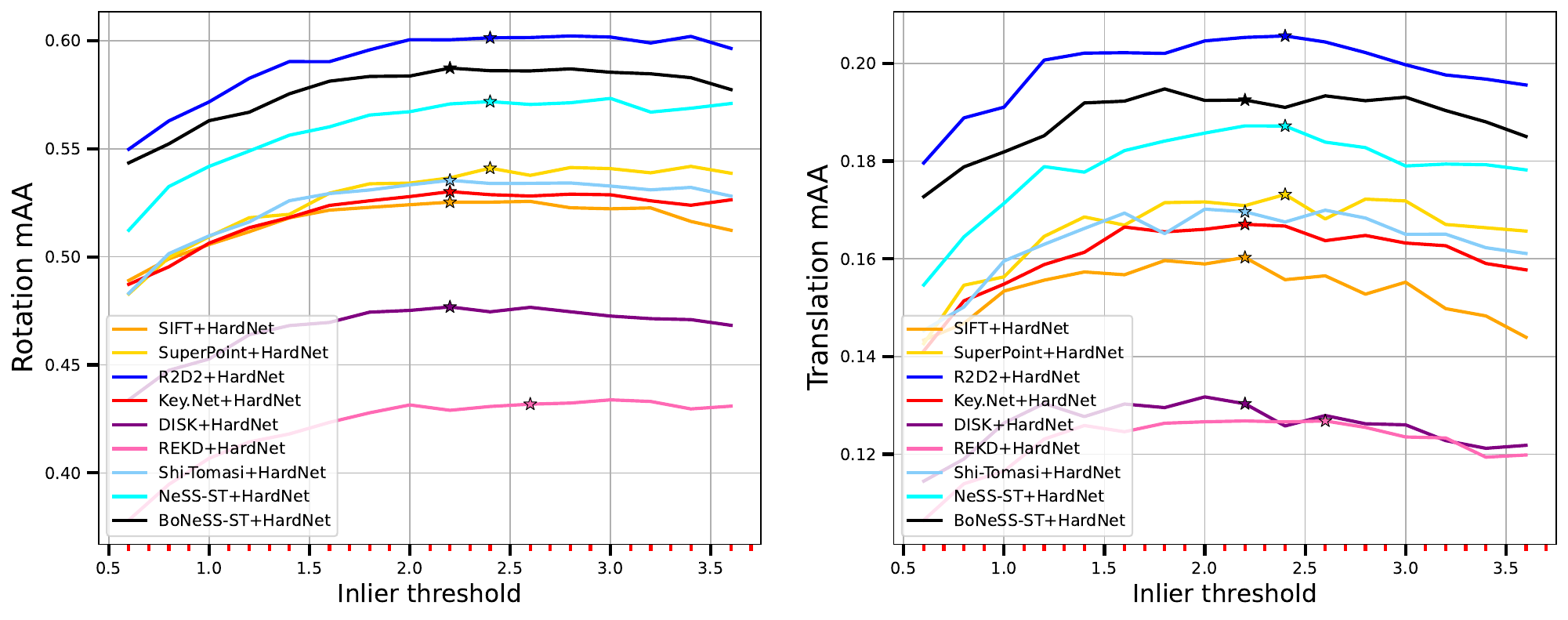}
    \caption{Inlier threshold tuning.}
    \end{center}
    \end{subfigure}
    \caption{\textbf{Initial tuning of method-specific hyperparameters for the ScanNet dataset.} We plot relative pose mAA-to-hyperparameter curves for rotation and translation calculated on the validation subset of ScanNet~\cite{dai2017scannet}. The star marks the best hyperparameter for a method according to the sum of rotation and translation mAA~\cite{yi2018learning, jin2021image} selected in a greedy manner~\cite{jin2021image}.}
    \label{fig:scannet_hyperparams_tuning}
    \end{center}
\end{figure}

\begin{figure}[h]
  \begin{center}
  \includegraphics[width=\linewidth]{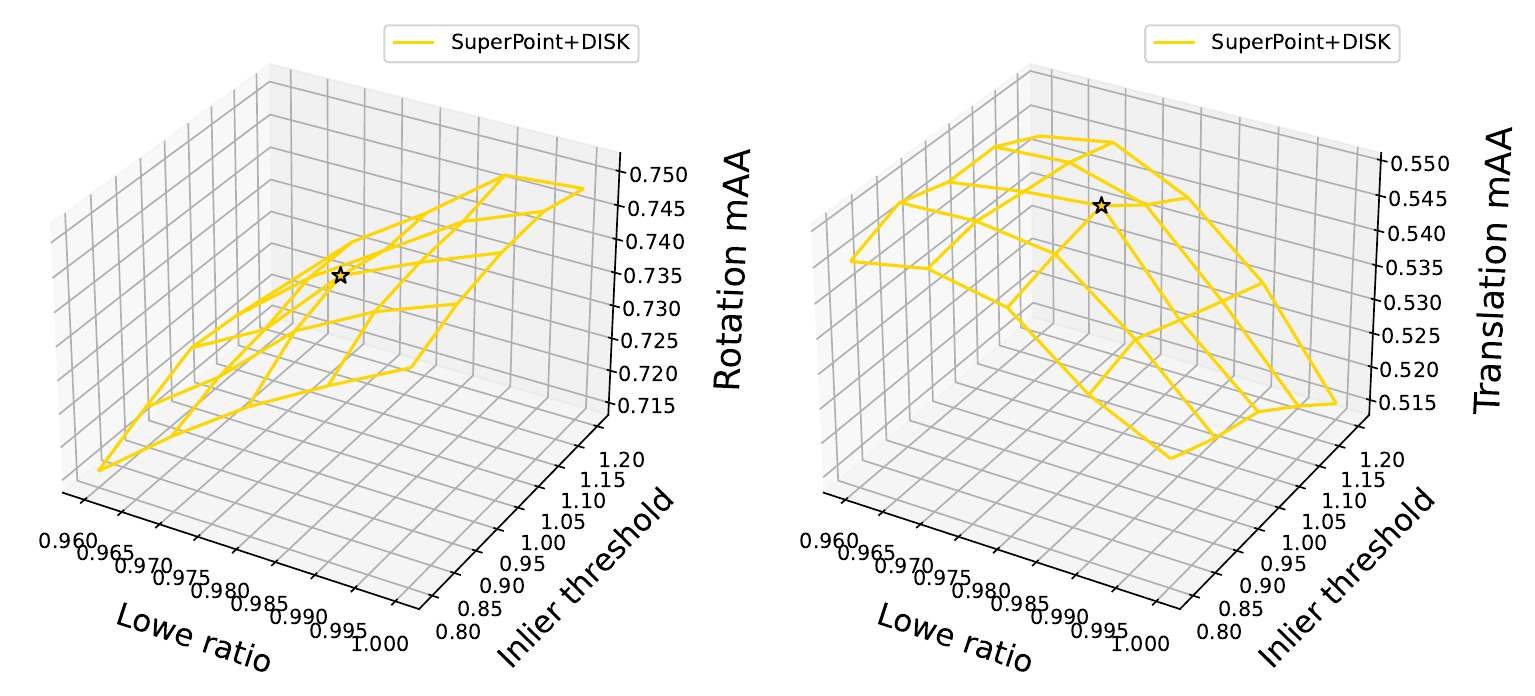}
  \end{center}
  \caption{\textbf{Grid search of optimal method-specific hyperparameters.} We plot relative pose mAA-to-hyperparameters grids for rotation and translation calculated on the validation subset of IMC-PT~\cite{jin2021image} for SuperPoint~\cite{detone2018superpoint}. The star marks the best pair of hyperparameters according to the sum of rotation and translation mAA~\cite{yi2018learning, jin2021image}.}
  \label{fig:grid_tuning_example}
\end{figure}

There are three important factors that significantly influence the outcomes of the evaluation on the two-view geometry estimation task: \begin{enumerate*}[label=(\roman*)] \item an image resolution~\cite{lowe2004distinctive}, \item a keypoint budget~\cite{jin2021image} and \item hyperparameters of two-view geometry estimators~\cite{jin2021image}.
\end{enumerate*} In our work, we evaluate all methods using images in the original resolution~\cite{pakulev2023ness} and choose the keypoint budget of 2048 keypoints as in~\cite{tyszkiewicz2020disk, jin2021image, pakulev2023ness, santellani2024gmm}. Note that we use different estimator hyperparameters such as the confidence level and the maximum number of RANSAC iterations for validation, tuning and ablations and testing. Using a high confidence level and a large iteration budget for testing ensures that the results reported in Table~\ref{tab:evaluation} can be repeated over multiple independent evaluation runs with minimal deviations. At the same time, using a more restrictive computational budget for validation, tuning and ablations allows us to keep the consumption of computational resources at a manageable level. To ensure a proper pick of the ratio test threshold and the inlier threshold for each method, we build on the hyperparameter tuning pipeline from~\cite{jin2021image, pakulev2023ness} and improve it with a grid search from~\cite{barath2023large}. Specifically, we begin by choosing an initial value for each hyperparameter via a greedy selection strategy~\cite{jin2021image} (see Fig.~\ref{fig:heb_hyperparams_tuning}, Fig.~\ref{fig:imc_pt_hyperparams_tuning} and Fig.~\ref{fig:scannet_hyperparams_tuning}). Then, we do a grid search in a small neighborhood around initial values as shown in Fig.~\ref{fig:grid_tuning_example}. We do not provide plots like in Fig.~\ref{fig:grid_tuning_example} for all methods on all datasets since that would take a lot of space. Instead, we summarize the final values of tuned hyperparameters in Table~\ref{tab:grid_tuning}. Refer to Appx.~\ref{appx:limitations:evaluation_protocol} for more details about the evaluation protocol.

Since in our work we evaluate the performance of keypoint detectors, we limit the usage of techniques that improve the performance of feature matching but are not related to keypoint detection. Concretely, we permit multi-scale keypoint detection for methods that support it: SIFT~\cite{lowe2004distinctive}, Key.Net~\cite{barroso2019key} and REKD~\cite{lee2022self}, as scale selection is at the core of the classical feature selection mechanism~\cite{lindeberg1998feature}. However, we do not allow to use the found scale to extract a better descriptor for a keypoint. This is to disambiguate the contribution of a better positioning of a feature point brought by a multi-scale detector, which is what a good keypoint detector should have, from the gains provided by the descriptor due to a better feature representation. For the same reason, we do not employ the orientation estimation module for SIFT~\cite{lowe2004distinctive} and REKD~\cite{lee2022self}, which use it, respectively, for extracting better descriptors and correspondence filtering. For each method, we use the reference implementation provided by the authors for inference and feature extraction. In some cases, we modify the reference implementation while keeping it as close to the original as possible to ensure the compatibility of a method with our evaluation pipeline and protocol (see Appx.~\ref{appx:limitations:evaluation_protocol} for more details).

\begin{table}[h]
  \caption{\textbf{Two-view geometry estimators and their hyperparameters.} We report the confidence level (Conf. level) and the maximum number of RANSAC iterations (Max. num. of iter.).}
  \label{tab:two_view_geometry_estimators}
  \begin{center}
  \resizebox{\columnwidth}{!}{
  \begin{tabular}{cccccc}
  \toprule

  \multirow{3}{*}{Estimator} & \multirow{3}{*}{Dataset} & \multicolumn{2}{c}{Validation / Tuning / Ablations} & \multicolumn{2}{c}{Testing} \\[2pt]
   & & \multirow{2}{*}{\makecell{Conf. \\ level}} & \multirow{2}{*}{\makecell{Max. num. \\ of iter.}} & \multirow{2}{*}{\makecell{Conf. \\ level}} & \multirow{2}{*}{\makecell{Max. num. \\ of iter.}} \\
  & & & & & \\

  \midrule

  \multirow{3}{*}{DEGENSAC~\cite{chum2005two}} & HEB~\cite{barath2023large} & \multirow{3}{*}{0.9999} & \multirow{3}{*}{200k} & \multirow{3}{*}{0.999999} & \multirow{3}{*}{200k} \\
   & IMC-PT~\cite{jin2021image} & & \\
   & MegaDepth~\cite{li2018megadepth} & & \\[4pt]
   
   OpenGV~\cite{kneip2014opengv} & ScanNet~\cite{dai2017scannet} & 0.9999 & 10k & 0.999999 & 200k \\

  \bottomrule
  \end{tabular}
  }
  \end{center}
\end{table}

\begin{table}[h]
  \caption{\textbf{Method-specific two-view geometry estimator hyperparameters.} We report the final values of hyperparameters tuned using the grid search strategy~\cite{barath2023large}: ratio test threshold and inlier threshold.}
  \label{tab:grid_tuning}
  \begin{center}
  \begin{tabular}{ccccccc}
  \toprule
  
  \multirow{3}{*}{Method} &
  \multicolumn{2}{c}{HEB~\cite{barath2023large}} &
  \multicolumn{2}{c}{IMC-PT~\cite{jin2021image}} &
  \multicolumn{2}{c}{ScanNet~\cite{dai2017scannet}} \\ 
  
  & \multirow{2}{*}{\makecell{Ratio test \\ threshold}} & \multirow{2}{*}{\makecell{Inlier \\ threshold}} & 
  \multirow{2}{*}{\makecell{Ratio test \\ threshold}} & \multirow{2}{*}{\makecell{Inlier \\ threshold}} &
  \multirow{2}{*}{\makecell{Ratio test \\ threshold}} & \multirow{2}{*}{\makecell{Inlier \\ threshold}} \\
  
  & & & & & & \\
  
  \midrule
  
  SIFT~\cite{lowe2004distinctive} & 
  1.0 & 2.6 & 
  0.99 & 0.7 &  
  0.96 & 2.2 \\
  
  SuperPoint~\cite{detone2018superpoint} &
  1.0 & 2.8 & 
  0.98 & 1.0 &  
  0.93 & 2.8 \\
  
  R2D2~\cite{revaud2019r2d2} & 
  1.0 & 2.6 & 
  0.99 & 1.0 &  
  0.92 & 2.4 \\
  
  Key.Net~\cite{barroso2019key} & 
  1.0 & 3.0 & 
  1.0 & 1.2 &  
  0.95 & 2.2 \\
  
  DISK~\cite{tyszkiewicz2020disk} & 1.0 & 2.6 & 
  0.98 & 0.9 &  
  0.93 & 2.2 \\
  
  REKD~\cite{lee2022self} & 
  1.0 & 3.0 & 
  0.98 & 1.4 &  
  0.91 & 2.4 \\
  
  \midrule
  
  Shi-Tomasi~\cite{shi1994good} & 
  1.0 & 2.2 & 
  0.99 & 0.7 &  
  0.93 & 2.4 \\
  
  NeSS-ST~\cite{pakulev2023ness} & 1.0 & 2.6 & 
  0.98 & 0.7 &  
  0.91 & 2.4 \\

  BoNeSS-ST &
  1.0 & 3.0 & 
  0.99 & 0.7 &  
  0.92 & 2.4 \\
  
  \bottomrule
  \end{tabular}
  \end{center}
\end{table}

\section{Limitations}
\label{appx:limitations}

\subsection{Stability Score}
\label{appx:limitations:stability_score}

In Fig.~\ref{fig:limitations:two_modes}, we demonstrate the key problem with the employed parameterization of $\beta$-EME: repeatability and EME are both quantified via the Euclidean distance. While this is not a problem for good keypoints, which can be approximated well by a unimodal distribution, or bad keypoints for which the distribution of projections has many modes, it is a problem for keypoints that have a few well-separated and localized modes. Table~\ref{tab:stability_score_limitations} shows that as the training progresses BoNeSS-ST improves its ability to predict the stability score for low-saliency keypoints, which greatly benefits its downstream performance on ScanNet. However, at the same time, it learns to avoid scoring high keypoints with multiple modes, which decreases the repeatability of detected keypoints and affects the performance on HEB, where the repeatability of keypoints matters more due to challenging viewpoint transformations between views. BoNeSS-ST trained for 10 epochs is the model we report in Table~\ref{tab:beta_ablation}, whereas BoNeSS-ST trained for 16 epochs is the model we report in Table~\ref{tab:evaluation}.

\begin{table}[h]
  \caption{\textbf{Study of limitations of the stability score on HEB and ScanNet datasets.} We report relative pose rotation (R) and translation (t) mAA~\cite{yi2018learning, jin2021image} for a $10\degree$ threshold as well as the number of inliers (NI) for BoNeSS-ST on validation subsets of HEB and ScanNet. The best results are in \first{red}.}
  \label{tab:stability_score_limitations}
  \centering
  \begin{tabular}{cccccc}
    \toprule
    \multirow{2}{*}{\makecell{Number of \\ training epochs}} &
    \multicolumn{2}{c}{HEB~\cite{barath2023large}} &
    \multicolumn{3}{c}{ScanNet~\cite{dai2017scannet}} \\
    
    & R & NI & 
    R & t & NI \\
    
    \midrule

    10 epochs &
    \first{0.233} & \first{89.7} & 
    0.579 & 0.193 & 128.1 \\

    16 epochs &
    0.231 & 81.4 & 
    \first{0.584} & \first{0.195} & \first{128.5} \\

    \bottomrule
  \end{tabular}
\end{table}

Inherently, the stability score is not a keypoint detection mechanism but a keypoint selection mechanism. Thus, it needs a provider of keypoints, a base detector, in order to be used in a detector, and, hence, the performance of a detector based on the stability score is always tied to the base detector. Fig.~\ref{fig:boness_st_and_shi_tomasi_1st_kp} illustrates this limitation of our approach. A keypoint ranked first by the Shi-Tomasi detector is not the best pick according to EME only because of its position. The distribution of keypoint projections indicates that the image structure it corresponds to actually has very good properties in terms of both repeatability and EME. In future work, we plan to address the aforementioned limitations of the stability score and the BoNeSS-ST detector.

\subsection{Evaluation Protocol}
\label{appx:limitations:evaluation_protocol}

Originally, in our experiments, we employed the pipeline from~\cite{pakulev2023ness}, which we had enhanced with a grid search of method-specific hyperparameters (see Appx.~\ref{appx:eval_protocol}). We report the results obtained using this pipeline in Table~\ref{tab:old_evaluation}. Note that the method-specific parameters used in this evaluation are the same as those reported in Table~\ref{tab:grid_tuning}. However, when reimplementing this evaluation pipeline to make it easy to use, we discovered that the results obtained on ScanNet are not stable. The issue was caused by the choice of the two-view geometry estimator hyperparameters, which we had taken from~\cite{pakulev2023ness}. These hyperparameters are marked in Table~\ref{tab:two_view_geometry_estimators} as the hyperparameters for validation, tuning and ablations. Therefore, we increased the confidence level and the number of RANSAC iterations for testing for all datasets to ensure that the outputs of the pipeline are consistent. Because tuning method-specific hyperparameters with these new hyperparameters on ScanNet is computationally demanding, we decided to use the old hyperparameters for validation, tuning and ablations. While the results on HEB, IMC-PT and MegaDepth did not change much by these changes, the results on ScanNet significantly improved for most methods. Using different two-view geometry estimator hyperparameters for tuning and testing can be regarded as a methodological flaw; however, we reason that this is acceptable since the hyperparameters of 
all methods listed in Table~\ref{tab:grid_tuning} are in a close range. Nevertheless, we plan to address this issue in future work. Also, we slightly changed the way that repeatability~\cite{schmid1998comparing} and MMA~\cite{mikolajczyk2005performance, dusmanu2019d2} are calculated to better match the definitions of the metrics.

Another change we introduced to the evaluation pipeline from~\cite{pakulev2023ness} concerns the top-k selection of keypoints. The original pipeline employs the same routine with non-maximum suppression and top-k score extraction for all methods. To ensure that each method performs in a way intended by the authors, we decided to use the code from a reference implementation of each method for top-k selection of feature points. This resulted in small improvements for SuperPoint~\cite{detone2018superpoint} and major improvements for REKD~\cite{lee2022self}. For Key.Net~\cite{barroso2019key}, it led to a substantial decrease in its downstream performance. Additionally, we switched from using our implementation of NeSS-ST, which is evaluated in Appx.~\ref{appx:ablation_study}, to the original implementation~\cite{pakulev2023ness} that decreased its performance. The results obtained using the original top-k selection routine~\cite{pakulev2023ness} and the original implementation of NeSS-ST are reported in Table~\ref{tab:old_evaluation}. This change is of importance because the optimal hyperparameters for the two-view geometry estimators, which we report in Table~\ref{tab:grid_tuning}, were tuned using the top-k selection routine from~\cite{pakulev2023ness}. We did not change the hyperparameters, as we did not have enough time to tune them again. While this is another methodological flaw of our evaluation, we believe that due to the reasons mentioned in the previous paragraph, the hyperparameters are close to their optimal values.

\begin{table}
  \caption{\textbf{Evaluation on the two-view geometry estimation task and with classical metrics using the previous version of the pipeline.} We report relative pose rotation (R) and translation (t) mAA~\cite{yi2018learning, jin2021image} for a $10\degree$ threshold as well as the number of inliers (NI). Also, we report repeatability (Rep.)~\cite{schmid1998comparing} and MMA~\cite{mikolajczyk2005performance, dusmanu2019d2} for a 3-pixel threshold. The top-3 results in each column are in \first{red}, \second{green} and \third{blue}.}
  \label{tab:old_evaluation}
  \begin{center}
  \resizebox{\columnwidth}{!}{
  \begin{tabular}{cccccccccccccc}
  \toprule

  \multirow{2}{*}{Method} &
  \multicolumn{2}{c}{HEB~\cite{barath2023large}} &
  \multicolumn{3}{c}{IMC-PT~\cite{jin2021image}} &
  \multicolumn{3}{c}{MegaDepth~\cite{li2018megadepth}} &
  \multicolumn{3}{c}{ScanNet~\cite{dai2017scannet}} &
  \multicolumn{2}{c}{HPatches~\cite{balntas2017hpatches}} \\ 
  
  & R & NI & 
  R & t & NI & 
  R & t & NI & 
  R & t & NI &
  Rep. & MMA \\
  
  \midrule

  SIFT~\cite{lowe2004distinctive} &
  0.182 & 49.2 & 
  0.713 & 0.387 & 181.2 & 
  0.845 & 0.295 & 383.6 & 
  0.576 & 0.195 & 106.9 &
  0.495 & 0.747 \\
  
  SuperPoint~\cite{detone2018superpoint} &
  0.225  & \third{55.7} & 
  0.709 & 0.370 & 114.5 &
  0.875 & 0.321 & 395.4 & 
  0.611 & 0.212 & 83.1 &
  \first{0.616} & \first{0.777} \\
  
  R2D2~\cite{revaud2019r2d2} &
  \second{0.232} & \second{65.9} &
  0.750 & 0.408 & \second{250.2} &
  0.873 & 0.313 & \second{452.0} &
  \second{0.643} & \first{0.235} & \first{166.3} &
  \third{0.549} & 0.755 \\
  
  Key.Net~\cite{barroso2019key} &
  0.202 & 54.4 &
  0.677 & 0.327 & 136.5 &
  0.848 & 0.276 & 384.4 &
  0.603 & 0.210 & 84.6 &
  \second{0.573} & 0.741 \\
  
  DISK~\cite{tyszkiewicz2020disk} &
  \first{0.249} & \first{127.5} & 
  \first{0.817} & \first{0.487} & \first{470.3} & 
  \second{0.884} & \third{0.323} & \first{622.3} &
  0.525 & 0.161 & 138.9 &
  0.547 & \second{0.768} \\
  
  REKD~\cite{lee2022self} &
  0.213 & \third{55.7} &
  0.605 & 0.269 & 128.2 & 
  0.856 & 0.271 & 396.1 & 
  0.453 & 0.139 & 141.5 &
  0.548 & 0.684 \\
  
  \midrule
  
  Shi-Tomasi~\cite{shi1994good} &
  0.196 & 57.0 & 
  0.751 & 0.426 & 215.0 & 
  0.864 & 0.311 & \third{429.0} & 
  0.568 & 0.194 & \third{148.9} &
  \second{0.573} & \first{0.777} \\
  
  NeSS-ST~\cite{pakulev2023ness} &
  0.229 & 44.0 & 
  \third{0.769} & \third{0.446} & 158.4 &
  \third{0.878} & \second{0.338} & 299.2 & 
  \second{0.621} & \third{0.218} & 106.8 &
  0.401 & 0.713 \\
  
  BoNeSS-ST &
  \third{0.230} & 54.7 & 
  \second{0.788} & \second{0.457} & \third{217.1} &
  \first{0.887} & \first{0.339} & 362.9 &
  \first{0.652} & \first{0.235} & \second{162.5} &
  0.476 & \third{0.757} \\
  
  \bottomrule
  \end{tabular}
  }
  \end{center}
\end{table}

Finally, we address not using multi-scale descriptor extraction and neural matchers in our evaluation. As can be seen from the discussions above, the contribution of keypoints to the performance is challenging to assess properly. Adding more elements to the pipeline will further aggravate the issue and make assessing the contribution of keypoints harder. Specifically, neural matchers are trained for a specific combination of a detector and a descriptor, so we need to provide weights trained or at least fine-tuned for each combination. Granted, our current evaluation of using a single descriptor for all methods is not perfect. Ideally, we want to select or even train a descriptor for each method. Concerning the multi-scale descriptor extraction, to have it in our pipeline, we need to implement it for methods that do not support it out of the box, e.g. DISK~\cite{tyszkiewicz2020disk}, as well as ensure that all methods use the same number of layers and the scale factors for constructing an image pyramid. While all the issues mentioned above stand, we believe that our pipeline employs the state-of-the-art methodology, so we leave them for future work.

\section{Additional Experiments}
\label{appx:additional_experiments}

\subsection{Evaluation on the Two-View Geometry Estimation Task}
\label{appx:two_view_geometry_eval}

\begin{figure}[h]
  \begin{center}
  \includegraphics[width=\linewidth]{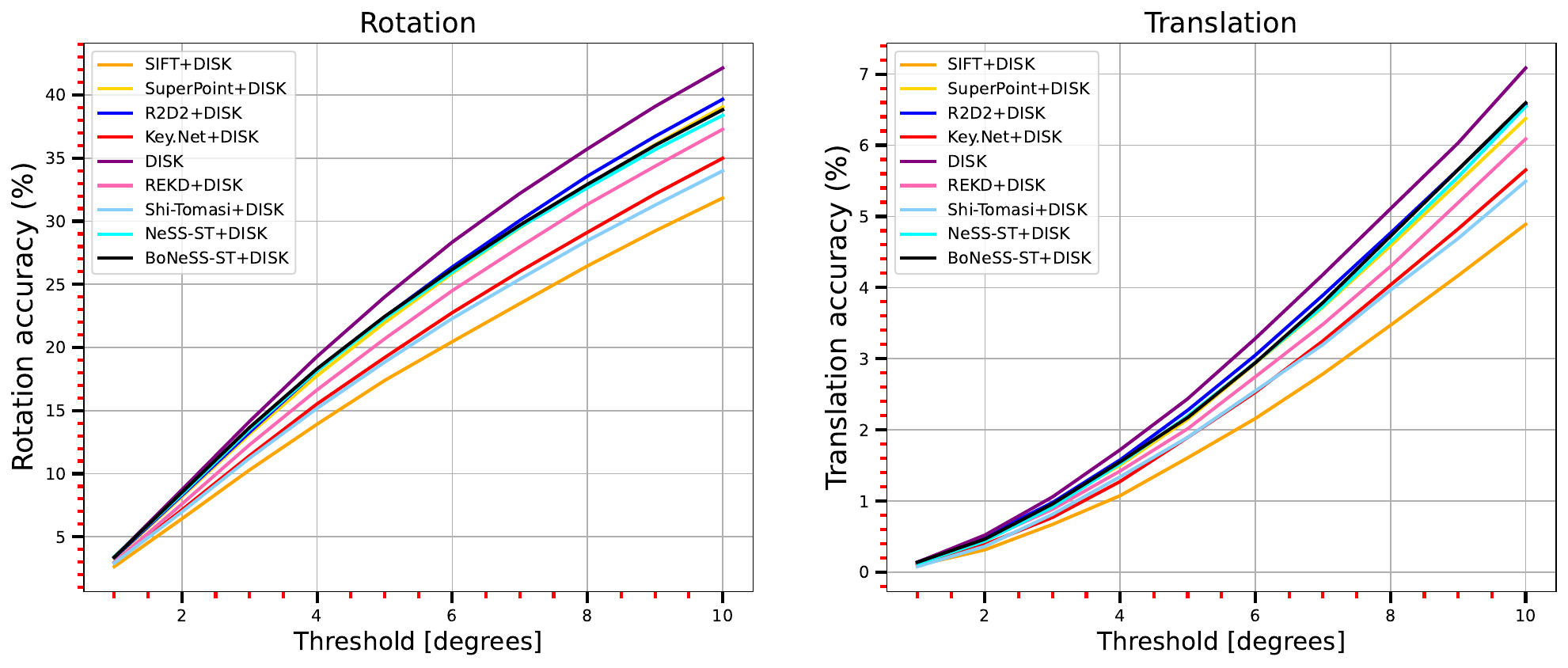}
  \end{center}
  \caption{\textbf{Evaluation on the HEB dataset on the planar homography estimation task.} We plot relative pose estimation accuracy-to-threshold curves for rotation and translation~\cite{yi2018learning, jin2021image} calculated on the test subset of HEB~\cite{barath2023large}.}
  \label{fig:heb_relative_pose_accuracy_to_threshold}
\end{figure}

\begin{figure}[h]
  \begin{center}
  \includegraphics[width=\linewidth]{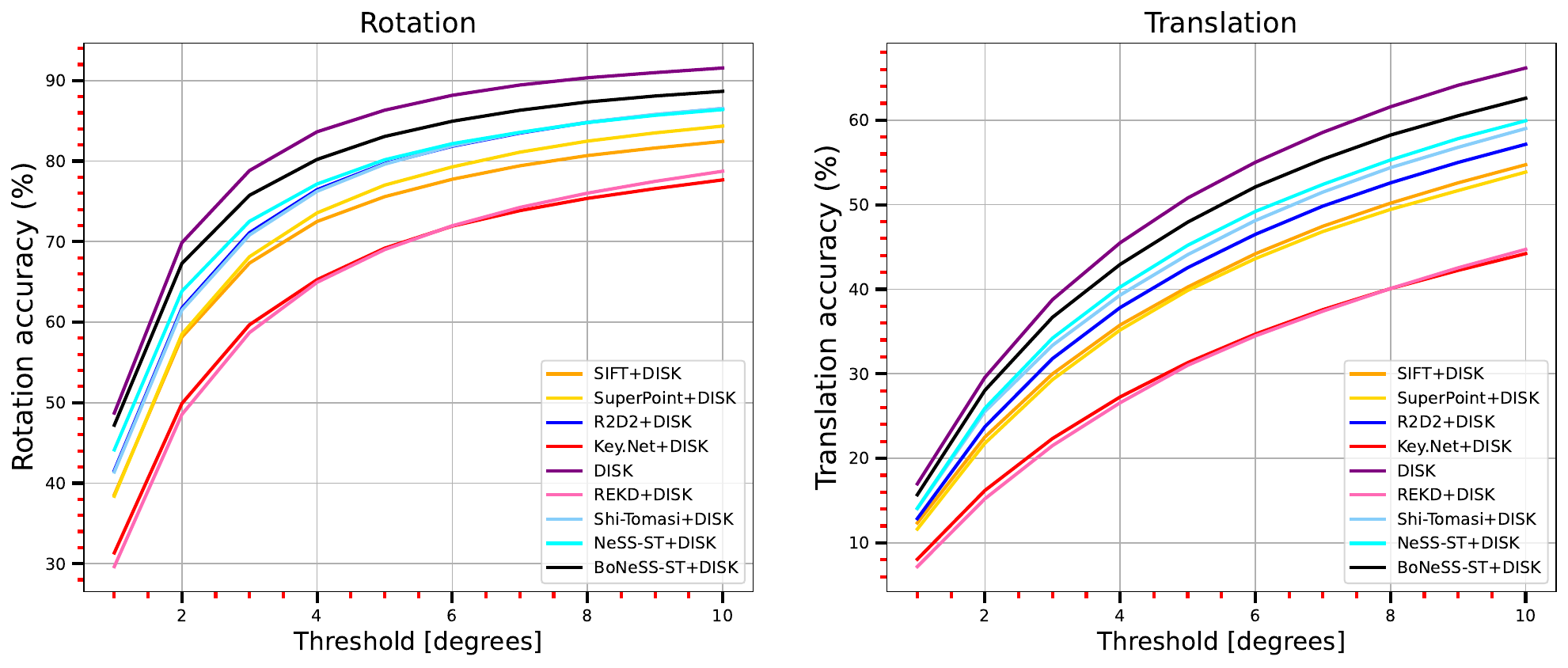}
  \end{center}
  \caption{\textbf{Evaluation on the IMC-PT dataset on the fundamental matrix estimation task.} We plot relative pose estimation accuracy-to-threshold curves for rotation and translation~\cite{yi2018learning, jin2021image} calculated on the test subset of IMC-PT~\cite{jin2021image}.}
  \label{fig:imc_pt_relative_pose_accuracy_to_threshold}
\end{figure}

\begin{figure}[h]
  \begin{center}
  \includegraphics[width=\linewidth]{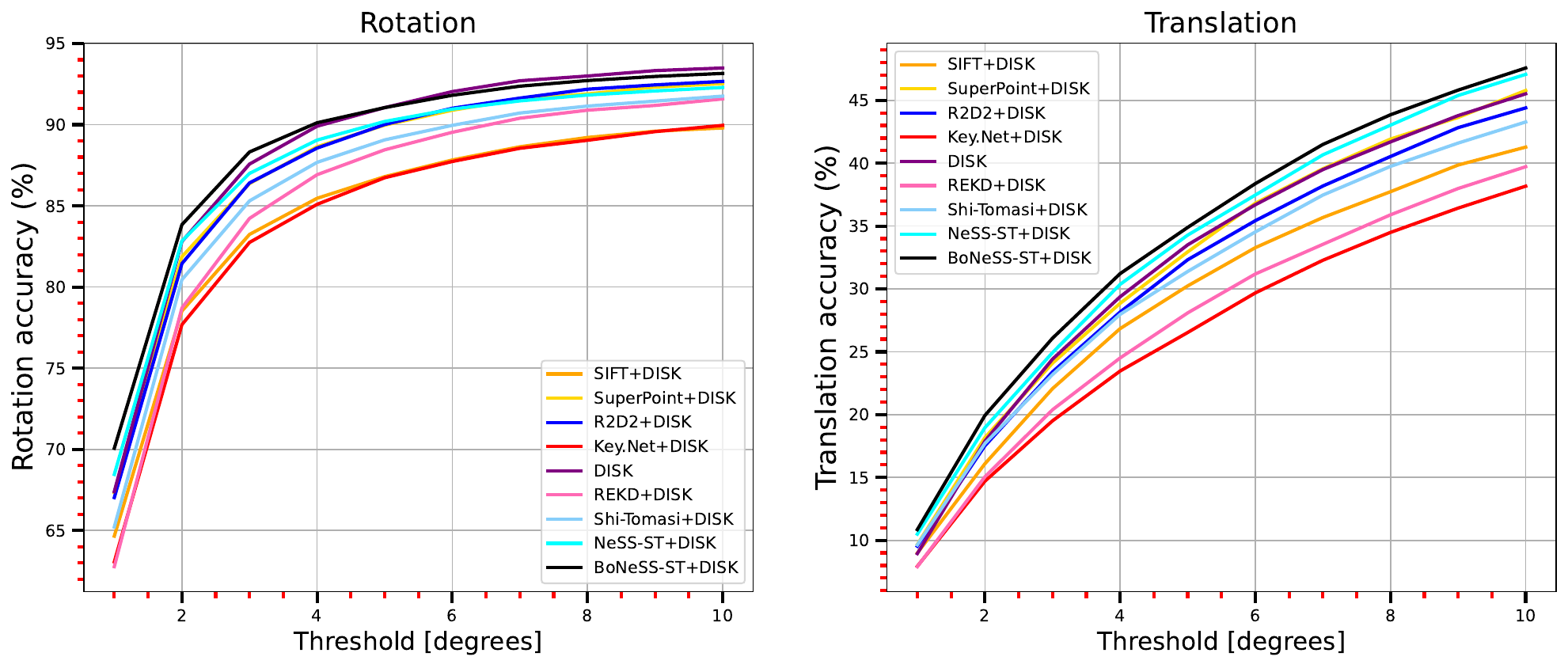}
  \end{center}
  \caption{\textbf{Evaluation on the MegaDepth dataset on the fundamental matrix estimation task.} We plot relative pose estimation accuracy-to-threshold curves for rotation and translation~\cite{yi2018learning, jin2021image} calculated on the test subset of MegaDepth~\cite{li2018megadepth}.}
  \label{fig:megadepth_relative_pose_accuracy_to_threshold}
\end{figure}

\begin{figure}[h]
  \begin{center}
  \includegraphics[width=\linewidth]{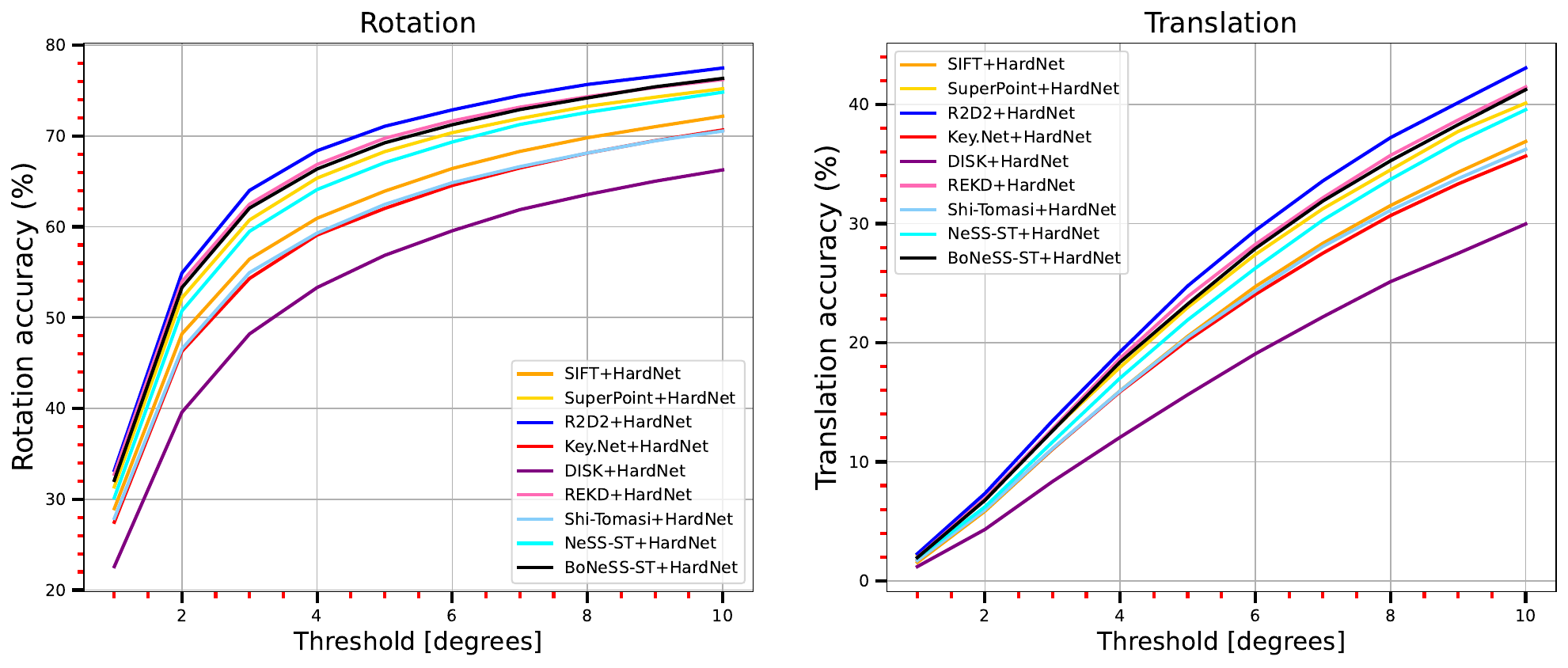}
  \end{center}
  \caption{\textbf{Evaluation on the ScanNet dataset on the essential matrix estimation task.} We plot relative pose estimation accuracy-to-threshold curves for rotation and translation~\cite{yi2018learning, jin2021image} calculated on the test subset of ScanNet~\cite{dai2017scannet}.}
  \label{fig:scannet_relative_pose_accuracy_to_threshold}
\end{figure}

In Fig.~\ref{fig:heb_relative_pose_accuracy_to_threshold}, Fig.~\ref{fig:imc_pt_relative_pose_accuracy_to_threshold}, Fig.~\ref{fig:megadepth_relative_pose_accuracy_to_threshold} and Fig.~\ref{fig:scannet_relative_pose_accuracy_to_threshold}, we provide relaitve pose accuracy-to-threshold curves~\cite{yi2018learning, jin2021image} based on which we calculate mAA (see Table~\ref{tab:evaluation}).

\subsection{Evaluation on Classical Metrics}
\label{appx:classical_metrics_eval}

\begin{figure}[h]
  \begin{center}
  \begin{subfigure}{\linewidth}
  \begin{center}
  \includegraphics[width=\linewidth]{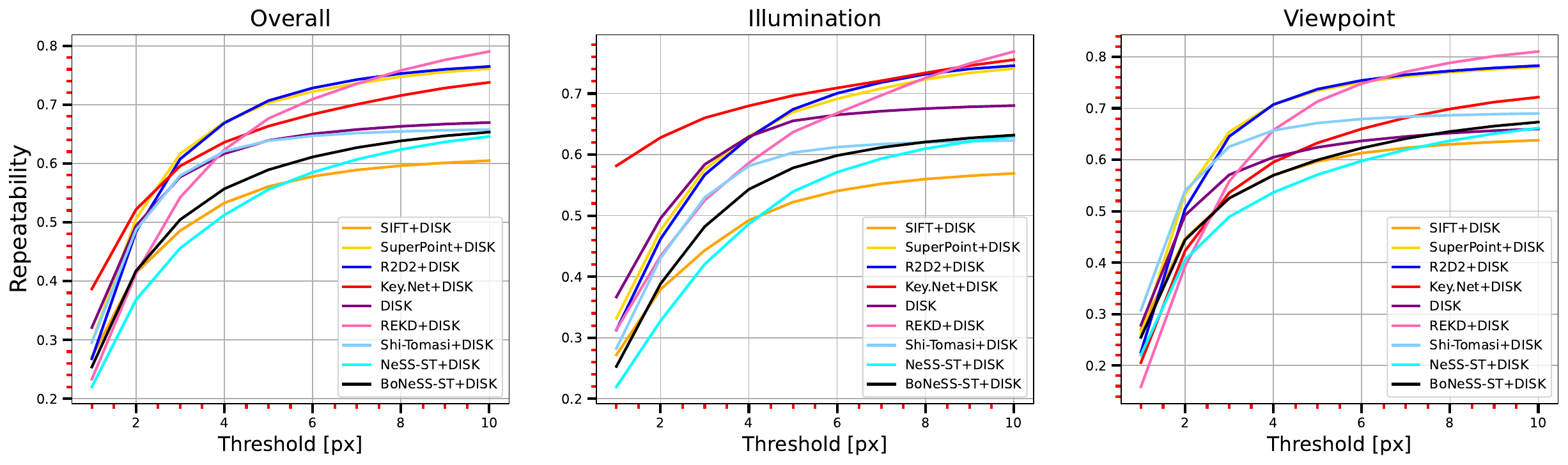}
    \caption{Repeatability-to-threshold curves.}
    \label{fig:hpatches_rep_curves}
  \end{center}
  \end{subfigure}
  \begin{subfigure}{\linewidth}
  \begin{center}
  \includegraphics[width=\linewidth]{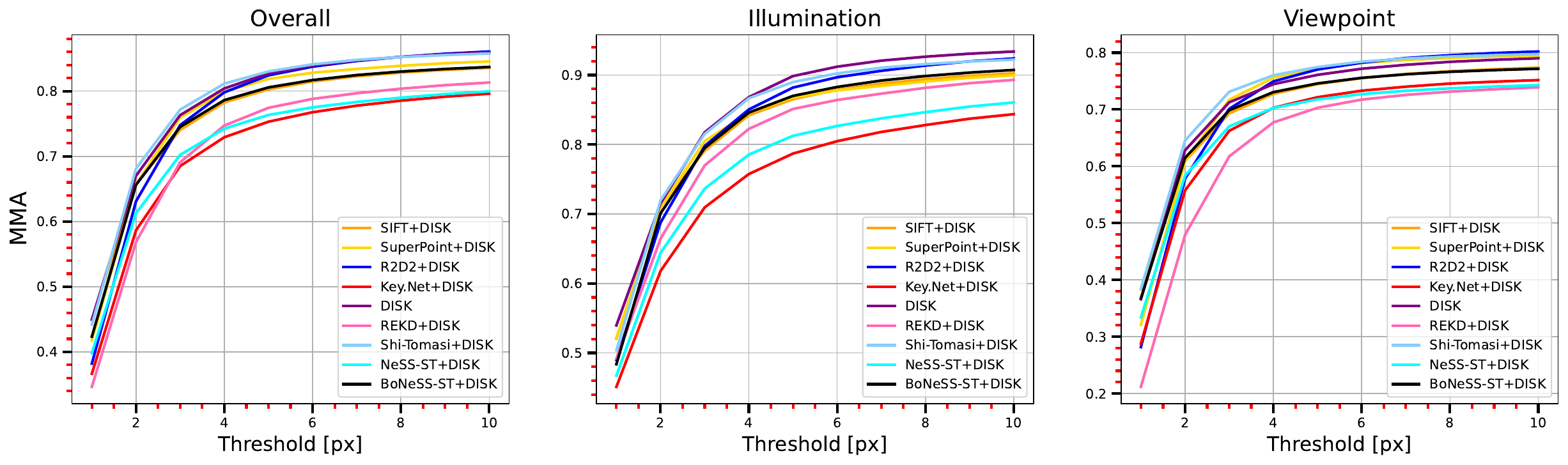}
  \caption{MMA-to-threshold curves.}
  \label{fig:hpatches_mma_curves}
  \end{center}
  \end{subfigure}
  \end{center}
  \caption{\textbf{Evaluation on the HPatches dataset on classical metrics.} We plot repeatability~\cite{schmid1998comparing}-to-threshold and MMA~\cite{mikolajczyk2005performance, dusmanu2019d2}-to-threshold curves calculated on the test subset of HPatches~\cite{balntas2017hpatches,dusmanu2019d2}.}
  \label{fig:hpatches_curves}
\end{figure}

In Table~\ref{tab:hpatches}, we provide a detailization of the results reported in Table~\ref{tab:evaluation} for illumination and viewpoint sequences of the HPatches~\cite{balntas2017hpatches} dataset. Additionally, in Fig.~\ref{fig:hpatches_curves}, we show repeatability~\cite{schmid1998comparing} and MMA~\cite{mikolajczyk2005performance, dusmanu2019d2} curves for all pixel thresholds.

\begin{table}
    \caption{\textbf{Evaluation on the HPatches dataset on classical metrics.} We report repeatability (Rep.)~\cite{schmid1998comparing} and MMA~\cite{mikolajczyk2005performance, dusmanu2019d2} for a 3-pixel threshold calculated on the test subset of HPatches~\cite{balntas2017hpatches}. We provide a detailization over illumination and viewpoint sequences. The top-3 results in each column are in \first{red}, \second{green} and \third{blue}.}
    \label{tab:hpatches}
    \begin{center}
    \begin{tabular}{ccccccc}
    \toprule
    
    \multirow{2}{*}{Method} &
    \multicolumn{2}{c}{Overall} &
    \multicolumn{2}{c}{Illumination} &
    \multicolumn{2}{c}{Viewpoint} \\ 
    
    & Rep. & MMA & 
    Rep. & MMA &  
    Rep. & MMA\\
    
    \midrule

    SIFT~\cite{lowe2004distinctive} &
    0.486 & 0.740 & 
    0.443 & 0.791 &
    0.525 & 0.693  \\
    
    SuperPoint~\cite{detone2018superpoint} &
    \first{0.617} & \third{0.759} & 
    \third{0.576} & \third{0.805} &
    \first{0.654} & \second{0.717}  \\
    
    R2D2~\cite{revaud2019r2d2} &
    \second{0.608} & 0.748 & 
    0.567 & 0.798 &
    \second{0.645} & 0.701  \\
    
    Key.Net~\cite{barroso2019key} &
    \third{0.596} & 0.685 & 
    \first{0.660} & 0.709 &
    0.536 & 0.662  \\
    
    DISK~\cite{tyszkiewicz2020disk} &
    0.577 & \second{0.763} & 
    \second{0.584} & \first{0.818} &
    0.571 & \third{0.713}  \\
    
    REKD~\cite{lee2022self} &
    0.542 & 0.691 & 
    0.525 & 0.770 &
    0.558 & 0.617 \\
    
    \midrule
    
    Shi-Tomasi~\cite{shi1994good} &
    0.579 & \first{0.772} & 
    0.530 & \second{0.815} &
    \third{0.625} & \first{0.731}  \\
    
    NeSS-ST~\cite{pakulev2023ness} &
    0.456 & 0.702 & 
    0.420 & 0.736 &
    0.489 & 0.670  \\
    
    BoNeSS-ST &
    0.504 & 0.745 & 
    0.482 & 0.795 & 
    0.525 & 0.699 \\
    
    \bottomrule
    \end{tabular}
    \end{center}
\end{table}

\subsection{Ablation Study}
\label{appx:ablation_study}

The design of the BoNeSS-ST detector is a result of studying how \begin{enumerate*}[label=(\roman*)]
\item different bounds (see Sec.~\ref{sec:boss}), 
\item measurement procedures (see Sec.~\ref{sec:syss}) and
\item supervisory signal generation procedures (see Sec.~\ref{sec:boness}) \end{enumerate*} influence the downstream performance. For this ablation, models were trained using the homography generation procedure from~\cite{pakulev2023ness} with $\beta=2.0$ and our implementation of the Shi-Tomasi detector, which is an improved version of the one from~\cite{pakulev2023ness}.

\paragraph{Results.} Table~\ref{tab:design_selection} shows that the improvements to the supervisory signal generation procedure, which are $\tnoise$ (see Sec.~\ref{sec:boness}) and Eq.~\ref{eq:shi_tomasi_measurement}, enhance the downstream performance when applied separately and together. Detector variants utilizing $\E[\|\rvkbari - k_i\|_2]$ and $\sqrt{\tr(\sigmabari) + \deltabari^{\top}\deltabari}$ with the aforementioned improvements have equal performance on IMC-PT~\cite{jin2021image}. For this reason, we additionally compared these models on ScanNet~\cite{dai2017scannet} (see Table~\ref{tab:design_selection_scannet}). In the evaluation, the model trained using $\sqrt{\tr(\sigmabari) + \deltabari^{\top}\deltabari}$ demonstrates the best overall performance, so we use the bound in Eq.~\ref{eq:eme_bound_jensen_sqrt_x2} in the final version of our detector. Table~\ref{tab:design_ablation} further demonstrates the validity of the adopted design choices on the test set of IMC-PT~\cite{jin2021image}. Note that we did not use the results from Table~\ref{tab:design_ablation} for the selection of the best-performing combination. The evaluation on the test set is meant solely to support our claims in Sec.~\ref{sec:syss} and Sec.~\ref{sec:boness} by evaluating with more data, which is crucial for having faithful mAA estimates~\cite{jin2021image}. The NeSS-ST detector ($\|\sigmabari\|_2$ with $\tsalient$) in Table~\ref{tab:design_ablation} demonstrates a better performance than NeSS-ST in Table~\ref{tab:evaluation} due to our enhanced implementation of the Shi-Tomasi detector. Similarly, the BoNeSS-ST detector in Table~\ref{tab:evaluation} has a better performance over the corresponding detector in Table~\ref{tab:design_ablation} ($\sqrt{\tr(\sigmabari) + \deltabari^{\top}\deltabari}$ with $\tsalient$, Eq.~\ref{eq:shi_tomasi_measurement} and $\tnoise$) due to the improved homography generation procedure (see Appx.~\ref{appx:generation_of_synth_views}).

\begin{table}[h]
  \caption{\textbf{Selection of the best combination of design choices.} We report relative pose rotation (R) and translation (t) mAA~\cite{yi2018learning, jin2021image} for a $10\degree$ threshold on the validation subset of IMC-PT~\cite{jin2021image}. The best results are in \first{red}. The best performing combination is selected according to a total sum of rotation and translation mAA.}
  \label{tab:design_selection}
  \begin{center}
  \resizebox{\columnwidth}{!}{
  \begin{tabular}{ccccccccc}
  \toprule
  \multirow{2}{*}{$\beta$-EME} & 
  \multicolumn{2}{c}{$\tsalient$~\cite{pakulev2023ness}} & 
  
  \multicolumn{2}{c}{$\tsalient$~\cite{pakulev2023ness} + Eq.~\ref{eq:shi_tomasi_measurement}} & 
  
  \multicolumn{2}{c}{$\tsalient$~\cite{pakulev2023ness} + $\tnoise$} & 
  
  \multicolumn{2}{c}{$\tsalient$~\cite{pakulev2023ness} + Eq.~\ref{eq:shi_tomasi_measurement} + $\tnoise$} \\
  
  & 
  R & t  & 
  R & t & 
  R & t & 
  R & t \\
  
  \midrule
  
  $\|\sigmabari\|_2$~\cite{pakulev2023ness} & 
  0.795 & 0.632 &
  0.801 & 0.633 & 
  0.806 & 0.648 & 
  0.807 & 0.649 \\
  
  $\tr(\sigmabari) + \deltabari^{\top}\deltabari$ &
  0.800 & 0.635 & 
  0.796 & 0.638 &
  0.806 & 0.649 &
  0.806 & 0.645 \\
  
  $\sqrt{\tr(\sigmabari) + \deltabari^{\top}\deltabari}$ & 
  0.788 & 0.633 &
  0.796 & 0.635 &
  0.805 & 0.648 &
  \first{0.811} & \first{0.648} \\
  
  $\E[\|\rvkbari - k_i\|_2]$ &
  0.796 & 0.633 & 
  0.799 & 0.640 & 
  0.809 & 0.646 & 
  \first{0.811} & \first{0.648} \\
  
  \bottomrule
  \end{tabular}
  }
  \end{center}
\end{table}

\begin{table}[h]
  \caption{\textbf{Selection of the best combination of design choices.} We report rotation (R) and translation (t) mAA~\cite{yi2018learning, jin2021image} for a $10\degree$ threshold on the validation subset of ScanNet~\cite{dai2017scannet}. The best results are in \first{red}. The best performing combination is selected according to a total sum of rotation and translation mAA.}
  \label{tab:design_selection_scannet}
  \begin{center}
  \begin{tabular}{ccc}
  \toprule
  \multirow{2}{*}{$\beta$-EME} &  
  \multicolumn{2}{c}{$\tsalient$~\cite{pakulev2023ness} + Eq.~\ref{eq:shi_tomasi_measurement} + $\tnoise$} \\
  
  & 
  R & t \\
  
  \midrule
  
  $\sqrt{\tr(\sigmabari) + \deltabari^{\top}\deltabari}$ &
  \first{0.586} & \first{0.198} \\

  $\E[\|\rvkbari - k_i\|_2]$ &
  0.567 & 0.187 \\

  \bottomrule
  \end{tabular}
  \end{center}
\end{table}

\begin{table}
  \caption{\textbf{Ablation study of the influence of various design choices on the performance.} We report relative pose rotation (R) and translation (t) mAA~\cite{yi2018learning, jin2021image} for a $10\degree$ threshold as well as the number of inliers (NI) on IMC-PT~\cite{jin2021image}. The best results are in \first{red}. The best performing combination for columns R and t is selected according to a total sum of rotation and translation mAA.}
    \label{tab:design_ablation}
    \begin{center}
    \resizebox{\linewidth}{!}{
    \begin{tabular}{ccccccccccccc}
    \toprule
    \multirow{2}{*}{$\beta$-EME} & 
    \multicolumn{3}{c}{$\tsalient$~\cite{pakulev2023ness}} & 
    
    \multicolumn{3}{c}{$\tsalient$~\cite{pakulev2023ness} + Eq.~\ref{eq:shi_tomasi_measurement}} & 
    
    \multicolumn{3}{c}{$\tsalient$~\cite{pakulev2023ness} + $\tnoise$} & 
    
    \multicolumn{3}{c}{$\tsalient$~\cite{pakulev2023ness} + Eq.~\ref{eq:shi_tomasi_measurement} + $\tnoise$} \\
    
    & 
    R & t & NI & 
    R & t & NI & 
    R & t & NI & 
    R & t & NI \\
    
    \midrule
    
    $\|\sigmabari\|_2$~\cite{pakulev2023ness} & 
    0.778 & 0.451 & 174.2 & 
    \first{0.781} & \first{0.453} & \first{191.7} & 
    0.780 & 0.456 & 206.2 & 
    0.780 & 0.459 & 208.9 \\
    
    $\tr(\sigmabari) + \deltabari^{\top}\deltabari$ &
    \first{0.782} & \first{0.454} & \first{187.8} & 
    \first{0.777} & \first{0.457} & 185.9 & 
    0.783 & 0.454 & 216.1 & 
    0.785 & 0.455 & 212.5 \\
    
    $\sqrt{\tr(\sigmabari) + \deltabari^{\top}\deltabari}$ & 
    0.765 & 0.451 & 169.8 & 
    0.776 & 0.455 & 179.4 & 
    0.780 & 0.460 & 202.3 & 
    \first{0.788} & \first{0.457} & \first{217.1} \\
    
    $\E[\|\rvkbari - k_i\|_2]$ &
    0.777 & 0.450 & 187.1 & 
    0.777 & 0.455 & 183.8 & 
    \first{0.785} & \first{0.456} & \first{206.3} & 
    \first{0.787} & \first{0.458} & 215.2 \\
    
    \bottomrule
    \end{tabular}
    }
    \end{center}
\end{table}

We extend the results of studying the influence of $\beta$ on the downstream performance from Sec.~\ref{sec:intrinsic_property} and elaborate on choosing the optimal $\beta$ for BoNeSS-ST. For $\beta$ abaltion, NeSS-ST models presented in Sec.~\ref{sec:intrinsic_property} were trained similarly to the models from the design ablation but utilize the implementation of Shi-Tomasi from~\cite{pakulev2023ness} during inference. Hence, their performance is slightly better than the performance of NeSS-ST in Table~\ref{tab:evaluation}, but it is inferior to the models from the design ablation. BoNeSS-ST models were trained in the same way as explained in Appx.~\ref{appx:training_details} expect the training time. 

\paragraph{Results.} Table~\ref{tab:beta_tuning} shows that the best-performing model has $\beta=2.828$.\footnote{The models with $\beta=2.378$ and $\beta=2.828$ have the same total mAA; however, when R and t are rounded separately, the model with $\beta=2.378$ has a better performance. To avoid the confusion in our reasoning behind the choice of the best-performing model, we modified the value of the translation mAA in Table~\ref{tab:beta_tuning} from $0.644$ to $0.643$.} We trained this model for 6 more epochs and obtained the reference BoNeSS-ST model, which is evaluated in Table~\ref{tab:evaluation}. By comparing the results in Table~\ref{tab:evaluation} and Table~\ref{tab:beta_ablation}, it can be seen that the reference model and the ablation model with $\beta=2.828$ have similar performance on the test set of IMC-PT~\cite{jin2021image}. However, the reference model has noticeably worse classical metrics. While this once again confirms the main claims of the paper, it also highlights an important limitaiton of the stability score discussed in Appx.~\ref{appx:limitations:stability_score}. Additionally, the results in Table~\ref{tab:beta_ablation} confirm that the predictions of our theoretical model are valid for NeSS-ST as well. Concretely, they show that $\beta$ is able to control the repeatability of keypoints detected by NeSS-ST, and the downstream performance of the detector depends on $\beta$ in a way that is described by our theoretical model. Similarly to the abalation study of the BoNeSS-ST design, we use the test set of IMC-PT in this ablation study solely for supporting our claims in Sec.~\ref{sec:intrinsic_property} with more data.

\begin{table}[h]
  \caption{\textbf{Tuning \bm{$\beta$} for BoNeSS-ST.} We report relative pose rotation (R) and translation (t) mAA~\cite{yi2018learning, jin2021image} for a $10\degree$ threshold on the validation subset of IMC-PT~\cite{jin2021image}. The best results are in \first{red}. The best performing $\beta$ is selected according to a total sum of rotation and translation mAA. When two models have the same performance, the model with the highest $\beta$ is selected.}
  \label{tab:beta_tuning}
  \centering
  \begin{tabular}{ccccccccc}
      \toprule
      $\beta$ & 1.189 & 1.414 & 1.681 & 2.000 & 2.378 & 2.828 & 3.363 & 4.0 \\ 
      \midrule
      R & 0.799 & 0.801 & 0.806 & 0.808 & \first{0.809} & \first{0.810} & 0.807 & 0.805 \\
      t & 0.632 & 0.634 & 0.639 & 0.642 & \first{0.643} & \first{0.642} & 0.641 & 0.638 \\
      \bottomrule
  \end{tabular}
\end{table}

\subsection{Qualitative Results}
\label{appx:qualitative_results}

\begin{figure}[h]
  \captionsetup[subfigure]{justification=centering}
  \begin{center}
    \begin{subfigure}{0.48\linewidth}
      \begin{center}
        \includegraphics[width=\linewidth]{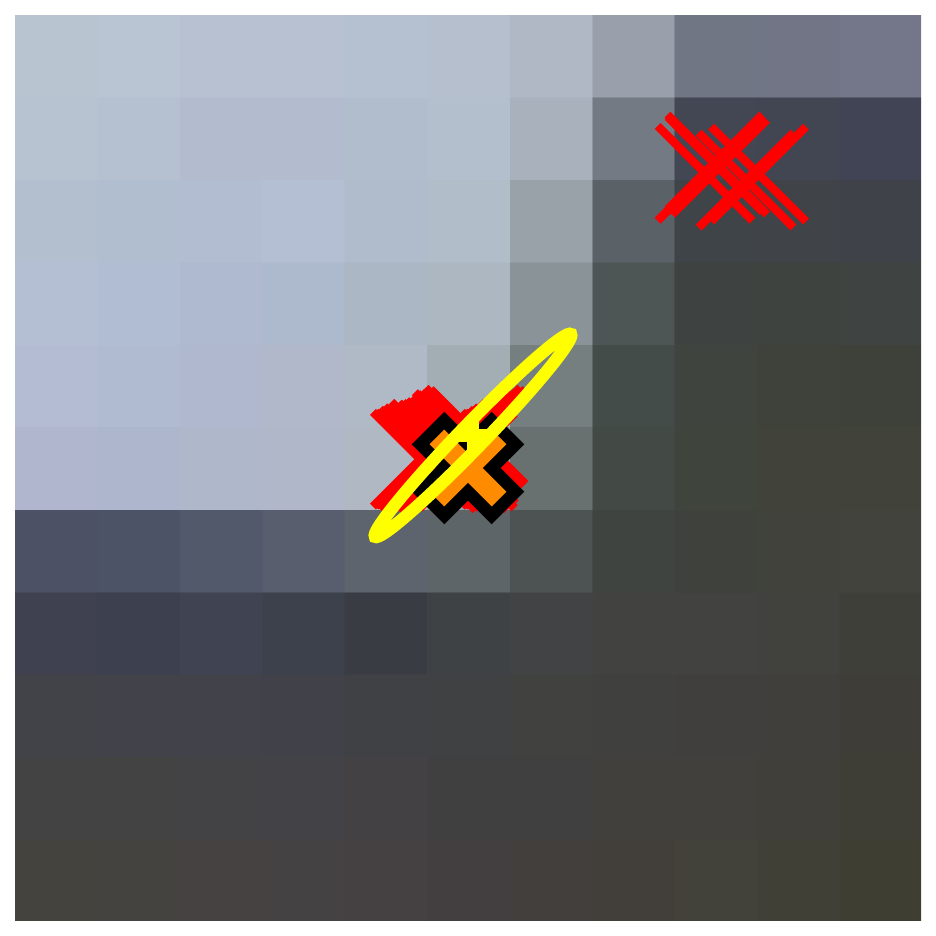}
      \end{center}
      \caption{A distribution of keypoint projections with two modes.}
      \label{fig:limitations:two_modes}
    \end{subfigure}
    \hfill
    \begin{subfigure}{0.48\linewidth}
      \begin{center}
        \includegraphics[width=\linewidth]{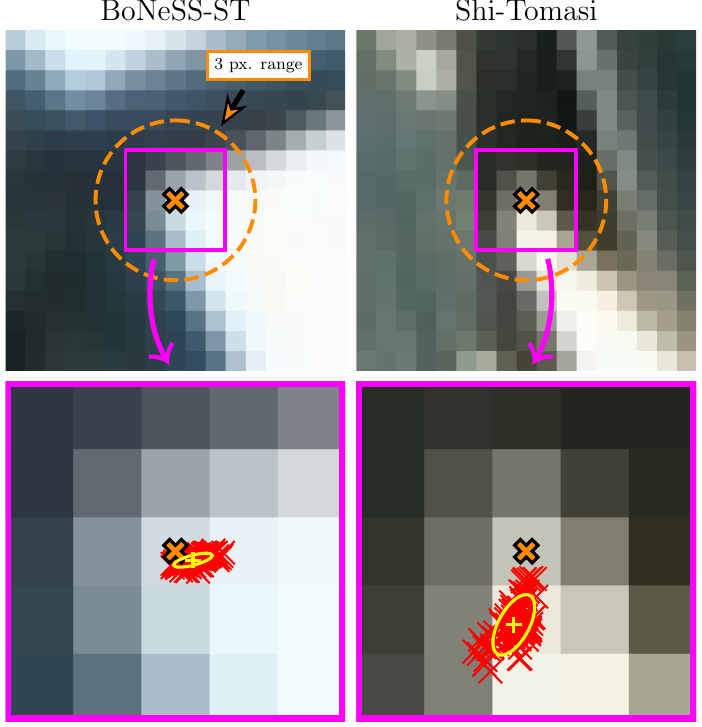}
      \end{center}
      \caption{The best keypoints according to BoNeSS-ST and Shi-Tomasi.}
      \label{fig:boness_st_and_shi_tomasi_1st_kp}
    \end{subfigure}
    \caption{\textbf{\subref{fig:limitations:two_modes}} A repeatable keypoint with a good EME has a salient structure in its vicinity. Since the structure is located at a considerable distance, it strongly affects $\beta$-EME of the keypoint. \textbf{\subref{fig:boness_st_and_shi_tomasi_1st_kp}} While both keypoints are repeatable within the 3-pixel threshold, they have a noticeably different quality in terms of $\beta$-EME.}
  \end{center}
\end{figure}

In Figure~\ref{fig:boness_st_and_shi_tomasi_1st_kp}, we highlight the difference in the scoring strategies of BoNeSS-ST and Shi-Tomasi detectors. 

\end{document}